\documentclass{article}

    \PassOptionsToPackage{numbers, compress}{natbib}

\usepackage[preprint]{neurips_2021}

\usepackage[utf8]{inputenc} 
\usepackage[T1]{fontenc}    
\usepackage[bookmarks=false]{hyperref}       
\usepackage{url}            
\usepackage{booktabs}       
\usepackage{amsfonts}       
\usepackage{nicefrac}       
\usepackage{microtype}      
\usepackage{xcolor}         

\usepackage{wrapfig}
\usepackage{times}
\usepackage{epsfig}
\usepackage{graphicx}
\usepackage{amsmath}
\usepackage{bbm}
\usepackage{multirow}
\usepackage{bigstrut}
\usepackage{booktabs}
\usepackage{subfig}
\usepackage{colortbl}
\usepackage{enumerate}
\usepackage{sidecap}
\usepackage{algorithm}
\usepackage{algpseudocode}
\usepackage{enumitem}
\usepackage{tablefootnote}

\title{Energy Aligning for Biased Models}

\author{%
  Bowen Zhao\thanks{work done by bowen during internship at tencent.} \\
  Tsinghua University\\
  \texttt{zbw18@mails.tsinghua.edu.cn} \\
  \And
  Chen Chen \\
  Tencent TEG AI \\
  \texttt{beckhamchen@tencent.com} \\
  \AND
  Qi Ju \\
  Tencent TEG AI \\
  \texttt{damonju@tencent.com} \\
  \And
  ShuTao Xia \\
  Tsinghua University, Peng Cheng Laboratory\\
  \texttt{xiast@sz.tsinghua.edu.cn} \\
}

\begin{document}

\maketitle

\begin{abstract}
Training on class-imbalanced data usually results in biased models that tend to predict samples into the majority classes, which is a common and notorious problem. From the perspective of energy-based model, we demonstrate that the free energies of categories are aligned with the label distribution theoretically, thus the energies of different classes are expected to be close to each other when aiming for ``balanced'' performance. However, we discover a severe energy-bias phenomenon in the models trained on class-imbalanced dataset. To eliminate the bias, we propose a simple and effective method named Energy Aligning by merely adding the calculated shift scalars onto the output logits during inference, which does not require to (i) modify the network architectures, (ii) intervene the standard learning paradigm, (iii) perform two-stage training. The proposed algorithm is evaluated on two class imbalance-related tasks under various settings: class incremental learning and long-tailed recognition. Experimental results show that energy aligning can effectively alleviate class imbalance issue and outperform state-of-the-art methods on several benchmarks.
\end{abstract}

\section{Introduction}
\label{sec:intro}

Deep neural networks (DNNs) have achieved tremendous success on a variety of challenging tasks~\cite{He2015DeepRL,devlin-etal-2019-bert,lin2019tsm,ren2015faster,he2020momentum}. The standard DNNs are usually reliable and powerful when the training data is representative of the evaluation data, i.e., the joint distribution of the observed training data $p_o(x, y)$ is consistent with that of the test data $p_t(x, y)$ ($x$ represents the data variable, and $y$ stands for the label variable). However, the ideal condition $p_o(x,y) = p_t(x,y)$ is not always satisfied in real-world applications. The degradation of model performance in realistic scenario due to distribution shift~\cite{quinonero2009dataset,lipton2018detecting,long2017deep} has been considered to be a main obstacle in the application and development of DNNs. In various distribution shift problems, class imbalance~\cite{he2009learning,huang2016learning} in training data is a common and intractable issue, which can be typically formulated as $p_o(y) \neq  p_t(y)$~\cite{cao2019learning,japkowicz2002class,quinonero2009dataset}. The huge class imbalance will lead to a biased model, which shows a strong tendency towards the majority classes.

The classical methods for tackling the biased models include re-sampling and re-weighting. Re-sampling attempts to rebalance the training data by oversampling the minority classes and/or undersampling the majority classes, but may result in overfitting to the minority classes or underutilization of the precious training data~\cite{shen2016relay,buda2018systematic,he2009learning,japkowicz2002class}. Re-weighting assigns different weights to different classes or training samples, which may cause optimization difficulty~\cite{huang2016learning,wang2017learning,byrd2019effect,ren2018learning,zhou2005training}. Recent studies reveal that decoupling the training procedure into representation learning and classification learning brings about improved overall performance~\cite{DBLP:conf/iclr/KangXRYGFK20,zhou2020bbn,Zhao_2020_CVPR,tang2020long,menon2020long}. However, most of these methods~\cite{DBLP:conf/iclr/KangXRYGFK20,zhou2020bbn,Zhao_2020_CVPR} are intuitive and heuristic, lacking theoretical foundations.

This paper endeavours to rectify the biased models from the theoretical perspective of energy-based models (EBMs)~\cite{lecun2006tutorial}, leading to a simple and effective approach called Energy Aligning. EBMs assign a scalar energy to each configuration of the variables (e.g., the pair $(x,y)$) --- low energies for compatible configurations while high energies for incompatible configurations conventionally. In this work, we reveal that the free energy of a specific category is theoretically aligned with the probability density of the class. Therefore, under the goal of pursuing ``balanced'' performance, the free energies of different classes are potentially expected to be similar since the target label distribution $p_t(y)$ is actually uniform for each class. However, as to a model trained on imbalanced data, we discover that a severe bias is hidden in the free energies: namely the minority classes commonly possess higher free energies while the majority classes usually hold lower free energies. Consequently, we put forward an algorithm --- Energy Aligning (EA) to combat the bias. EA attempts to align the free energies of the minority classes to those of the majority classes, which is simply achieved by correcting the predicted logits with the theoretically calculated shift scalars. We delineate that energy aligning can be applied to a purely discriminative classification model, meaning that one can easily obtain a corrected model without modifying the network architectures or altering the standard training procedure. EA is evaluated on two tasks related to class imbalance: class incremental learning~\cite{li2017learning,rebuffi2017icarl} and long-tailed recognition~\cite{liu2019large,DBLP:conf/iclr/KangXRYGFK20}, and surpasses state-of-the-art methods with simple operations.

Our key contributions can be summarized as follows: (i) Different from previous work, we theoretically derive a straightforward solution to deal with class imbalance from the energy-based perspective. (ii) We interpret the bias in models trained on imbalanced data from the view of energy values. Starting from this observation, we propose energy aligning for correcting biased models, which is not only a simple but also a principled approach. (iii) We demonstrate the effectiveness of energy aligning on class incremental learning and long-tailed recognition tasks, from which the experimental results indicate that energy aligning achieves wide applications and significant performance without tedious hyper-parameter tuning.

\section{Energy-Based Models and Discriminative Deep Neural Networks}

\textbf{Energy-Based Models.} Considering a model with two sets of variables $x$ (e.g., images) and $y$ (e.g., labels) ($x \in \mathcal{X}$, $y \in \mathcal{Y}$, $\mathcal{X}$ and $\mathcal{Y}$ denote the whole space of $x$ and $y$, respectively), energy-based models (EBMs) capture compatibility by associating a non-probabilistic, scalar energy to each configuration of the variables~\cite{lecun2006tutorial}. Generally, EBMs build an energy function $E_{\theta}(x, y)$ that maps each configuration $(x, y)$ to an energy value, where $\theta$ is a trainable parameter set. Conventionally, the small energy values represent highly compatible configurations of the variables (when $y$ is suitable for $x$), while large energy values signify highly incompatible configurations of the variables (when $y$ is incorrect for $x$). For the decision-making tasks, the system only needs to map the correct answer with the lowest energy, regardless of the magnitude of the energy value. Nevertheless, the uncalibrated energies from separate EBMs are probably not commensurate, so that the energies are required to be normalized in the scenario that the model is going to be used in combination with others. The most common approach is utilizing the Boltzmann distribution (also called Gibbs distribution) to turn the collection of energy values into a probability density: 
\begin{equation}
p_{\theta}(y | x) = 
\frac{e^{- E_{\theta}(x, y)}}
{\int_{y' \in \mathcal{Y}} e^{- E_{\theta}(x, y')}}
.
\label{eq:gibbs_con}
\end{equation}

\textbf{Discriminative Deep Neural Networks.} In recent years, a classification task with $C$ classes is typically optimized with a discriminative deep neural network $f_{\theta}(x)$ where $\theta$ means the learnable parameter set, by which the input variable $x$ is mapped to $C$ scalar values, known as logits. Then, these logits are employed to parameterize a posterior probability distribution over $C$ classes via the softmax function:
\begin{equation}
  p_{\theta}(y | x) = 
  \frac{e^{ f_{\theta}(x)[y]}}
  {\sum_{y'=1}^C e^{ f_{\theta}(x)[y']}},
\label{eq:p_y_con_x}
\end{equation}
where $f_{\theta}(x)[y]$ represents the $y^{\text{th}}$ index of $f_{\theta}(x)$, i.e., the logit corresponding to the $y^{\text{th}}$ class.

\textbf{Connection.} Compared Eq.~(\ref{eq:p_y_con_x}) with Eq.~(\ref{eq:gibbs_con}), energy-based models are inherently connected with discriminative neural networks~\cite{grathwohl2019your,liu2020energy}. Without changing the parameterization of the neural network $f_{\theta}(\cdot)$, we can re-use the logits to represent an energy of the configuration $(x, y)$ as
\begin{equation}
E_{\theta}(x, y) = - f_{\theta}(x)[y].
\label{eq:energy_f}
\end{equation}
In fact, the negative logit $- f_{\theta}(x)[y]$ does reflect the degree of compatibility between $x$ and $y$ --- small value of $- f_{\theta}(x)[y]$ corresponds to a ``good'' configuration while large value of $- f_{\theta}(x)[y]$ corresponds to a ``bad'' one, which is exactly what energy-based models require.

\section{Energy Aligning for Biased Models}
\label{sec:es}

When learning on a class-imbalanced dataset, the trained model will be biased towards the majority classes~\cite{liu2019large,DBLP:conf/iclr/KangXRYGFK20}. In this work, we derive a novel algorithm to rectify the biased model from the perspective of energy-based models, and achieve state-of-the-art performance. 

\textbf{Bias in Energies.} Similar to Eq.~\eqref{eq:gibbs_con}, for EBMs, the joint distribution of $x$ and $y$ can be formulated by using the Boltzmann distribution again, which is expressed as
\begin{equation}
p_{\theta}(x, y) = \frac{e^{- E_{\theta}(x, y)}}
{Z_{\theta}}, \quad
Z_{\theta} = \int_{x' \in \mathcal{X}, y' \in \mathcal{Y}} e^{- E_{\theta}(x', y')}.
\label{eq:joint_d}
\end{equation}
Then, by marginalizing the joint distribution Eq.~\eqref{eq:joint_d} over $x$, we obtain
\begin{equation}
  p_{\theta}(y) 
  = \frac{ \int_{x' \in \mathcal{X}} e^{- E_{\theta}(x', y) }}{Z_{\theta}} 
  = \frac{ e^{- E_{\theta}(y) }}{Z_{\theta}},
\label{eq:p_y}
\end{equation}
where the Helmholtz free energy
$
  E_{\theta}(y) = -
  \log \int_{x' \in \mathcal{X}} e^{- E_{\theta}(x', y)}.
$
Then, by taking the logarithm of both sides of Eq.~\eqref{eq:p_y}, we obtain
\begin{equation}
  \log p_{\theta}(y) = - E_{\theta}(y) - \log Z_{\theta}.
\label{eq:log_p_y}
\end{equation}
As shown in Eq.~\eqref{eq:log_p_y}, the negative free energy $-E_{\theta}(y)$ is linearly aligned with $\log p_{\theta}(y)$, i.e., $-E_{\theta}(y) \propto \log p_{\theta}(y)$. When it comes to the discriminative neural network $f_{\theta}(\cdot)$, combined with Eq.~\eqref{eq:energy_f}, we obtain the negative free energy with model $f_{\theta}(\cdot)$:
$
- E_{\theta}(y) 
=
\log \int_{x' \in \mathcal{X}} e^{ f_{\theta}(x')[y]}
.
$

Due to class imbalance, the label distribution of the observed training set is different from that of the target (test) set, which are written as $p_o(y)$ and $p_t(y)$ respectively. Though the observed training data is imbalanced, it is still expected to achieve balanced predictions and superior overall results during test. In fact, in the popular and common evaluation protocols for class imbalance issue, models are evaluated on the corresponding balanced test data after learning on the imbalanced training data, which alludes a hidden condition that the target label distribution $p_t(y)$ is exactly uniform for all classes of interest. Ideally, the model parameterized with $\theta$ is expected to reflect the target label distribution, i.e., $p_{\theta}(y) \approx p_t(y)$. Therefore, it is desirable to have 
\begin{equation}
  p_{\theta}(y=i) = p_{\theta}(y=j), \quad 1 \leq i \leq C, \quad 1\leq j\leq C, \quad i \neq j.
\label{eq:p_t_should}
\end{equation}
Moreover, combined with Eq.~\eqref{eq:log_p_y}, we expect to have:
\begin{equation}
-E_{\theta}(y=i) = -E_{\theta}(y=j), \quad 1 \leq i \leq C, \quad 1\leq j\leq C, \quad i \neq j.
\label{eq:e_t_should}
\end{equation}

\begin{wrapfigure}[11]{r}{0.38\textwidth}
  \centering
  \vspace{-4mm}
  \subfloat[raw model]{\label{fig:energy_plot_a} \includegraphics[width=0.18\textwidth]{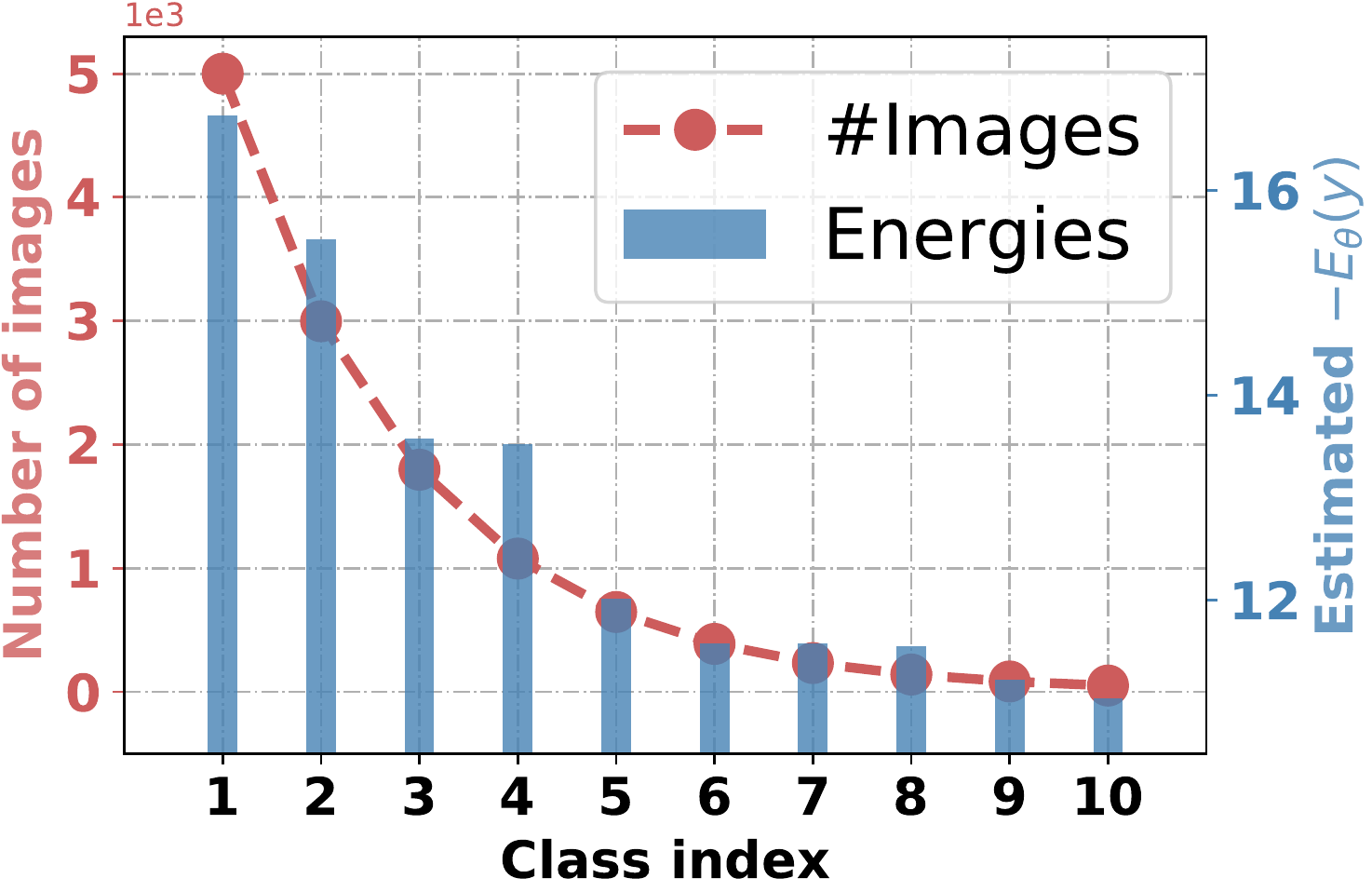}}
  \subfloat[corrected model]{\label{fig:energy_plot_b} \includegraphics[width=0.18\textwidth]{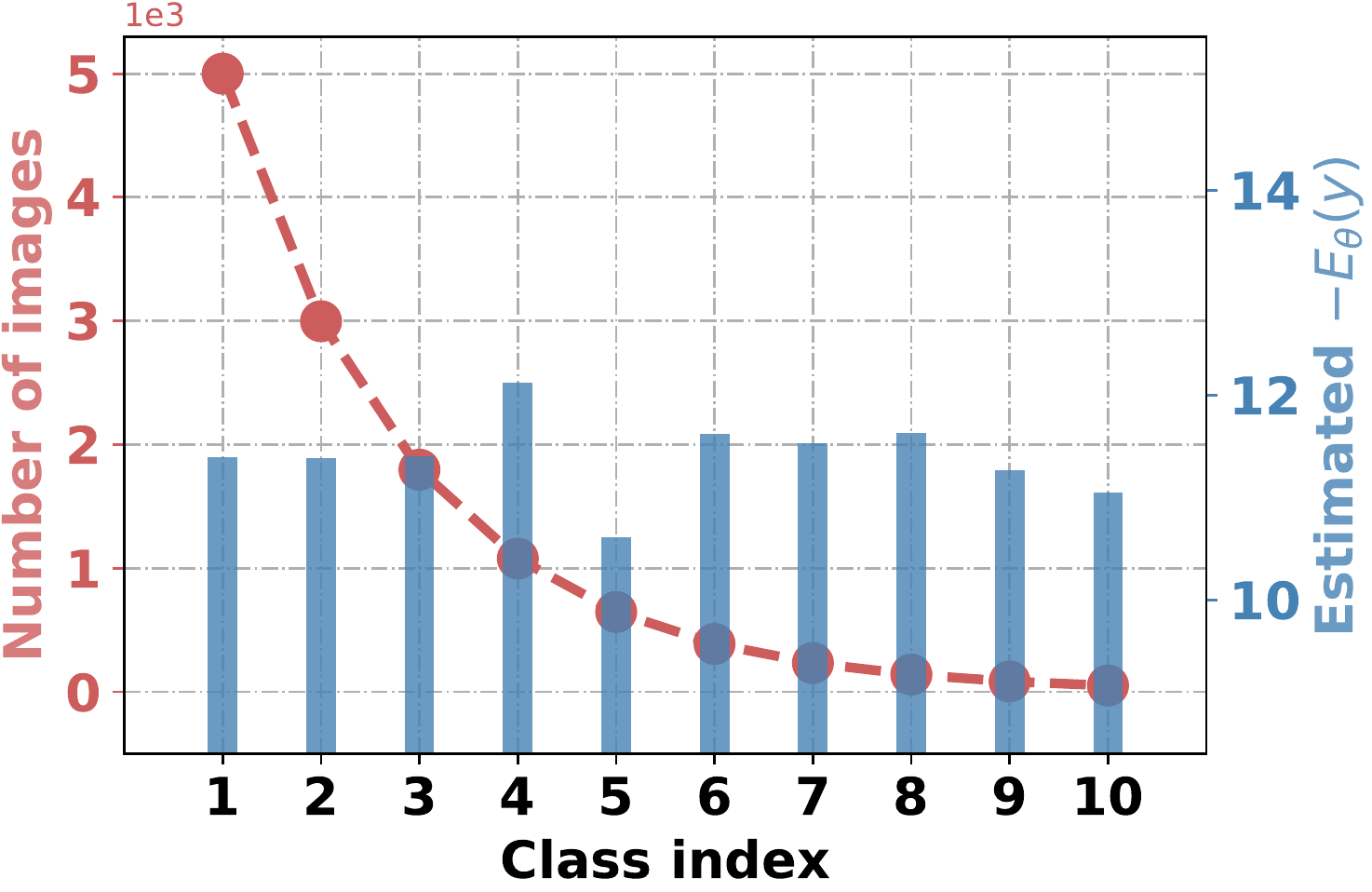}} 
  \caption{Number of training images per class and estimated negative free energy per class on CIFAR10-LT.}
  \label{fig:energy_plot}
\end{wrapfigure}
Unfortunately, the model trained on imbalanced data actually tends to manifest the label distribution of the training data set, i.e., $p_o(y)$ instead of $p_t(y)$. To verify the above hypothesis, we conduct an experiment on CIFAR10-LT (the class-imbalanced version of CIFAR10 with label distribution of its training data illustrated in Fig.~\ref{fig:energy_plot_a}). Based on the learned model, we estimate the free energies of the ten categories, bringing to light that the majority classes tend to have lower energies whereas the minority classes usually hold higher energies as shown in Fig.~\ref{fig:energy_plot_a}. This phenomenon does not conform to Eq.~\eqref{eq:e_t_should}, exposing the bias in the trained model from the perspective of energy-based learning.

\textbf{Energy Aligning.} The above analysis have pointed out that the prior label distribution $p_t(y)$ hidden in the goal is actually uniform for classes of interest. However, the free energies $E_{\theta}(y)$ of the model trained on imbalanced data does not match the prior knowledge (Eq.~\eqref{eq:e_t_should}). Instead of intervening the training process of $f_{\theta}(\cdot)$, we propose a simple and effective method, called Energy Aligning (EA), to correct the biased model directly. In EA, the logits are refined to make the energies satisfy Eq.~\eqref{eq:e_t_should}.

Without loss of generality, considering the free energies of the $i^{\text{th}}$ and $j^{\text{th}}$ class ($1 \leq i \leq C, 1\leq j\leq C, i \neq j$), we view the $i^{\text{th}}$ category as the ``anchor class'' and rebalance the two free energies with a shift scalar $\alpha_j$, which is formulated as
\begin{equation}
\begin{aligned}
& -E_{\theta}(y=i) = -E_{\theta}(y=j) + \alpha_j \\
\Rightarrow \quad& 
 \log \int_{x' \in \mathcal{X}} e^{ f_{\theta}(x')[i]}
= 
\alpha_j +  \log \int_{x' \in \mathcal{X}} e^{ f_{\theta}(x')[j]} \\
\Rightarrow \quad&
\alpha_j =  \log \frac{\int_{x' \in \mathcal{X}} e^{ f_{\theta}(x')[i]}}
{\int_{x' \in \mathcal{X}} e^{ f_{\theta}(x')[j]}}.
\label{eq:alpha}
\end{aligned}
\end{equation}
Assuming that the shift scalar $\alpha_j$ has already been calculated (will be described later), the right side of Eq.~\eqref{eq:alpha} can be reformulated as
\begin{equation}
\begin{aligned}
  -E_{\theta}(y=j) + \alpha_j 
  &= \alpha_j +  \log \int_{x' \in \mathcal{X}} e^{ f_{\theta}(x')[j]} \\
  &=  \log \int_{x' \in \mathcal{X}} e^{ f_{\theta}(x')[j] + \alpha_j} .
\end{aligned}
\label{eq:re_logits}
\end{equation}
As shown in Eq.~\eqref{eq:re_logits}, for any input data point, the $j^{\text{th}}$ logit can be adjusted by straightforwardly adding a non-learned scalar $\alpha_j$, leading to the advantage that the negative free energies $-E_{\theta}(y=i)$ and $-E_{\theta}(y=j)$ of the $i^{\text{th}}$ and $j^{\text{th}}$ classes could be equalized. Intuitively, when $\int_{x' \in \mathcal{X}} e^{ f_{\theta}(x')[i]} >
\int_{x' \in \mathcal{X}}  e^{ f_{\theta}(x')[j]}
$, $\alpha_j$ is positive, then the logit $f_{\theta}(x')[j]$ is augmented by $\alpha_j$; if $\int_{x' \in \mathcal{X}} e^{ f_{\theta}(x')[i]} <
\int_{x' \in \mathcal{X}}  e^{ f_{\theta}(x')[j]}
$, $\alpha_j$ is negative, then the logit $f_{\theta}(x')[j]$ is attenuated by $\alpha_j$. 

\begin{wrapfigure}[13]{r}{0.35\textwidth}
  \centering
  \includegraphics[width=0.33\textwidth]{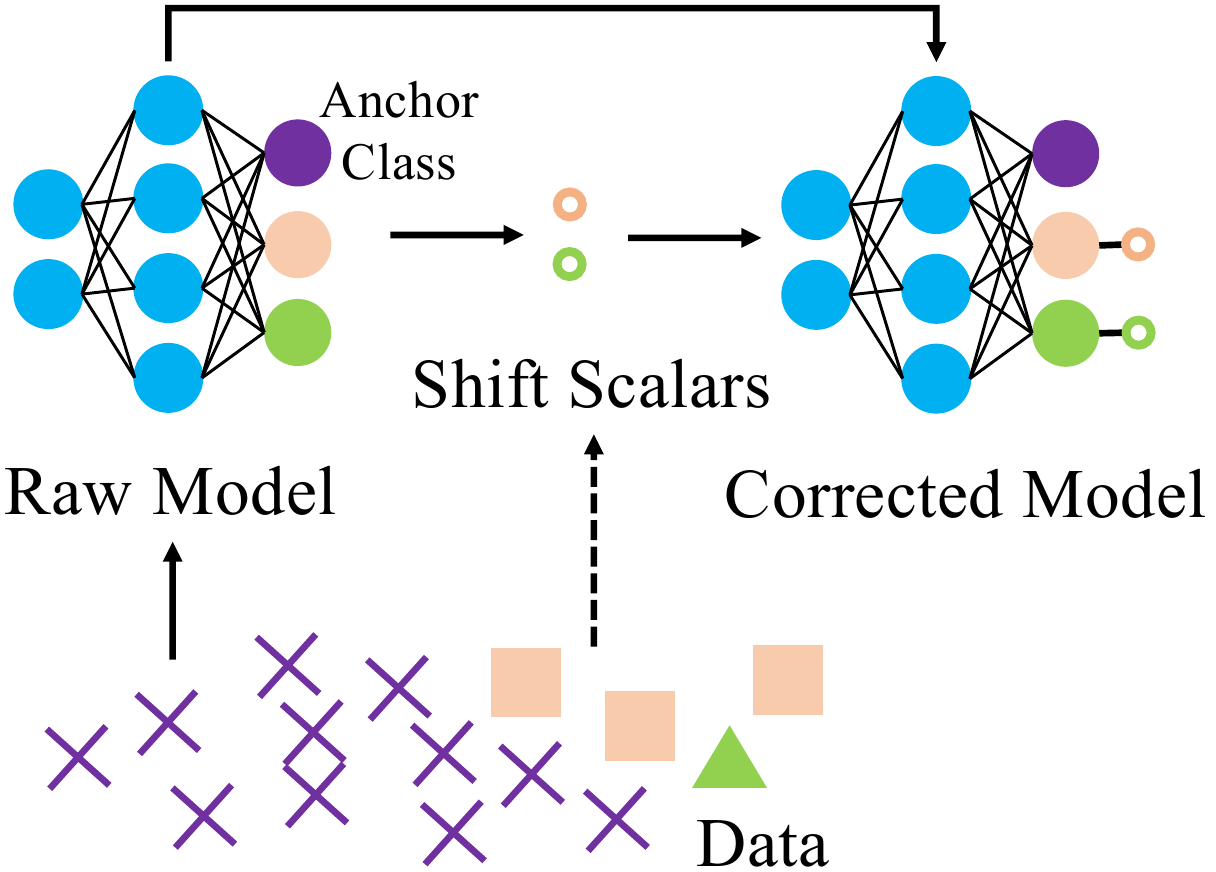} 
  \caption{Framework of energy aligning for biased models.}
  \label{fig:framework}
\end{wrapfigure}
In terms of any other category $j \ (j \neq i)$, we can adjust the $j^{\text{th}}$ logit with corresponding scalar $\alpha_j$ to satisfy Eq.~\eqref{eq:e_t_should} similarly, hence the model with energy aligning would treat all categories equally and make Eq.~\eqref{eq:p_t_should}, Eq.~\eqref{eq:e_t_should} hold. The corrected model is written as $f_{\theta;\{\alpha_j\}_{j=1}^{C-1}}$ here. The proposed energy aligning strategy is illustrated in Fig.~\ref{fig:framework}. As an example, for the experiment on CIFAR10-LT, the negative free energies of the corrected model is depicted in Fig.~\ref{fig:energy_plot_b}, which shows that energy aligning can indeed balance the energies.

\textbf{Approximation of Shift Scalar $\alpha_j$.} In order to acquire $\alpha_j$, the remaining essential problem is how to calculate the terms $\int_{x' \in \mathcal{X}} e^{ f_{\theta}(x')[i]}$ and $\int_{x' \in \mathcal{X}} e^{ f_{\theta}(x')[j]}$. Avoiding directly computing the intractable integrals, in this work, we employ Monte Carlo integration~\cite{caflisch1998monte,hammersley2013monte} to estimate the integrals. Specifically, for the $i^{\text{th}}$ category, the integral is approximated by 
\begin{equation*}
\int_{x' \in \mathcal{X}} e^{ f_{\theta}(x')[i]} 
\approx
\int_{x' \in \mathcal{X'}} e^{ f_{\theta}(x')[i]} 
\approx
\frac{1}{S} \sum_{s=1}^{S} \frac{e^{ f_{\theta}(x_s) [i]}}{q(x_s)},
\end{equation*}
where $\mathcal{X'} \subset \mathcal{X}$ is the input space of categories of interest. Since $e^{ f_{\theta}(x)[i]}$ with the trained $\theta$ usually outputs very low response to samples that are almost unrelated to the categories of interest, it suffices to estimate the integral on the subspace $\mathcal{X'}$. Data points $\{x_s\}_{s=1}^{S}$ are sampled from the proposal distribution $q(x)$, which is assumed to assign uniform values to samples from the input space $\mathcal{X'}$ (i.e., $q(x)$ is a uniform distribution). The same is true for another integral $\int_{x' \in \mathcal{X}} e^{ f_{\theta}(x')[j]}$. Then, we have
\begin{equation}
\begin{aligned}
  \alpha_j \approx  \log 
  \frac{
  \sum_{s=1}^{S} e^{ f_{\theta}(x_s)[i]} 
  }
  {
  \sum_{s=1}^{S} e^{ f_{\theta}(x_s)[j]} 
  }
  &=  
  LSE_{s=1}^{S}( f_{\theta}(x_s)[i]) -
  LSE_{s=1}^{S}( f_{\theta}(x_s)[j])
,
\end{aligned}
\label{eq:hat_alpha}
\end{equation}
where $LSE_{s=1}^{S}(z_s) = \log \sum_{s=1}^{S} e^{z_s}$. 

In practice, as to the chosen of samples $\{x_s\}^{S}_{s=1}$ related to categories of interest, two possible strategies are feasible. The first one is to prepare a balanced validation set on which $\alpha_j$ can be calculated. Another more data-saving option is to re-sample a balanced set from the training data (we adopt this strategy here). Different augmentations can be applied to obtain abundant samples that is beneficial for alleviating the estimation error caused by overfitting. Besides augmentations, to further mitigate the insufficient sample issue, we can divide the categories into $M$ clusters within which the same shift scalar is shared, based on the number of training samples per class. Specifically, for class clusters $\mathcal{O}_i$ and $\mathcal{O}_j$ with $C_i$ and $C_j$ categories respectively, similar to Eq.~\eqref{eq:alpha}, Eq.~\eqref{eq:re_logits} and Eq.~\eqref{eq:hat_alpha}, we view cluster $\mathcal{O}_i$ as the ``anchor cluster'' and rebalance the average energies for the two clusters with a shared shift scalar $\alpha_j$, which can be formulated as
\begin{equation*}
  \begin{aligned}
  & \frac{1}{C_i} \sum_{i \in \mathcal{O}_i} -E_{\theta}(y=i) = \alpha_j +\frac{1}{C_j} \sum_{j \in \mathcal{O}_j} -E_{\theta}(y=j)\\
  \Rightarrow \quad& 
  \alpha_j = 
    \frac{1}{C_i} \sum_{i\in \mathcal{O}_i} \log 
    \int_{x' \in \mathcal{X}} e^{ f_{\theta}(x')[i]} - \frac{1}{C_j} \sum_{j \in \mathcal{O}_j} \log \int_{x' \in \mathcal{X}} e^{ f_{\theta}(x')[j]}
    .
  \end{aligned}
\end{equation*}
And the estimation of $\alpha_j$ is given by
\begin{equation}
  \alpha_j \approx 
  \frac{1}{C_i} \sum_{i \in \mathcal{O}_i} LSE_{s=1}^S( f_{\theta}(x_s)[i]) -
  \frac{1}{C_j} \sum_{j \in \mathcal{O}_j} LSE_{s=1}^S( f_{\theta}(x_s)[j])
  .
  \label{eq:cluster_alpha}
\end{equation}
Moreover, from the below equation
\begin{equation}
\begin{aligned}
  \alpha_j +\frac{1}{C_j} \sum_{j \in \mathcal{O}_j} -E_{\theta}(y=j) = \frac{1}{ C_j} \sum_{j \in \mathcal{O}_j} \log \int_{x' \in \mathcal{X}} e^{ f_{\theta}(x')[j] + \alpha_j },
\label{eq:cluster_es}
\end{aligned}
\end{equation}
one can still adjust the logits of classes in cluster $\mathcal{O}_j$ by the shared $\alpha_j$ directly for any input data to obtain a corrected model, written as $f_{\theta;\{\alpha_j\}_{j=1}^{M-1}}$.

Even though more sophisticated integral approximation methods can be introduced to estimate the shift scalars, we demonstrate that surprising performance can be achieved by employing above simple strategies in comprehensive experiments described in the following sections.

\section{Application I: Class Incremental Learning}
\label{sec:ex_il}

In class incremental learning (CIL), models are expected to be able to learn new categories from streaming data without forgetting old categories that they have mastered. Unfortunately, DNNs undergo a serious problem known as catastrophic forgetting~\cite{French1999CatastrophicFI,MichaelCatastrophic}, which reveals that DNNs can hardly recollect old knowledge after learning new knowledge. Many approaches for catastrophic forgetting fail in this scenario~\cite{hsu2018re}, while the simple rehearsal strategy, which uses a very small amount of old data to complement training data, has been demonstrated to be effective in CIL~\cite{hsu2018re,DBLP:journals/corr/abs-1904-07734}. Nevertheless, the class imbalance problem that the number of old-classes data is much smaller than that of new-classes data is severe, which is proved to be a vital factor for catastrophic forgetting in CIL~\cite{hou2019learning,wu2019large,Zhao_2020_CVPR}. Even some previous studies~\cite{wu2019large,Zhao_2020_CVPR,Belouadah_2019_ICCV,belouadah2020scail} focus on the problem of class imbalance in CIL, most of these methods are intuitive and lack a fundamental theory. 

\begin{figure}[t]
\begin{table}[H]
\centering
\caption{Class incremental learning performance (top-5 / top-1 accuracy $_{\pm \text{std}}$ \%) on ImageNet100 with 10 incremental steps (10 classes per step). The best results are in bold.}
\resizebox{\textwidth}{25mm}{
\begin{tabular}{c|ccccccccccc}
\toprule
\#step & 1  & 2  & 3 & 4 & 5 &6  &7 & 8 & 9  & 10 & Avg \\
\midrule
LwF~\cite{li2017learning} & 99.2& 95.4 & 86.2 & 74.1  & 63.9  & 55.1  & 50.3 & 44.5 & 40.4  & 36.6  & 60.7  \\
iCaRL~\cite{rebuffi2017icarl} & 99.5  & 97.8 & 94.1 & 91.8  & 88.0  & 82.7 & 77.3 & 73.2 & 67.3 & 63.8 & 81.8  \\
EEIL~\cite{castro2018end} & 99.4  & \textbf{99.0} & \textbf{96.4} & 93.8 & 90.4 & 88.8 & 86.6 & 84.9 & 82.2 & 80.2 & 89.2  \\
BiC~\cite{wu2019large} & 98.4 & 96.2 & 94.0 & 92.9 & 91.1 & 89.4  & 88.1 & 86.5 & 85.4 & 84.4 & 89.8  \\
RPS~\cite{Rajasegaran2019RandomPS} & 99.4 & 97.4 & 94.2 & 92.6 & 89.4 & 86.2 & 83.7 & 82.1 & 79.5 & 74.0 & 86.6  \\
WA~\cite{Zhao_2020_CVPR}  & 98.8  & 96.8 & 94.5 & 93.1  & 90.5 & 89.9 & 88.8 & 88.0 & 86.2 & 84.1 & 90.2\\
& /88.4  & /84.1 & /79.5 & /75.6  & /69.3  & /66.8 & /64.8 & /62.7 & /59.5 & /57.3 & /68.8 \\
\midrule
EA (Ours)  & $\mathop{\text{99.1}}\limits_{\pm 0.3}$ & $\mathop{\text{97.5}}\limits_{\pm 0.5}$ & $\mathop{\text{95.5}}\limits_{\pm 0.4}$  & $\mathop{\textbf{94.0}}\limits_{\pm 0.4}$  & $\mathop{\textbf{91.3}}\limits_{\pm 0.2}$ & $\mathop{\textbf{90.7}}\limits_{\pm 0.3}$ &  $\mathop{\textbf{89.7}}\limits_{\pm 0.0}$ &  $\mathop{\textbf{88.7}}\limits_{\pm 0.3}$ &  $\mathop{\textbf{87.5}}\limits_{\pm 0.3}$  &  $\mathop{\textbf{86.3}}\limits_{\pm 0.1}$ & $\mathop{\textbf{91.2}}\limits_{\pm 0.2}$ \\
&  $\mathop{\text{/88.6}}\limits_{\pm 1.0}$ & $\mathop{\textbf{/85.0}}\limits_{\pm 0.6}$ & $\mathop{\textbf{/80.8}}\limits_{\pm 0.4}$  & $\mathop{\textbf{/76.5}}\limits_{\pm 0.3}$  & $\mathop{\textbf{/71.6}}\limits_{\pm 0.4}$  & $\mathop{\textbf{/69.8}}\limits_{\pm 0.5}$ & $\mathop{\textbf{/67.4}}\limits_{\pm 0.3}$ & $\mathop{\textbf{/65.5}}\limits_{\pm 0.7}$ & $\mathop{\textbf{/62.7}}\limits_{\pm 0.3}$  & $\mathop{\textbf{/60.3}}\limits_{\pm 0.5}$ & $\mathop{\textbf{/71.1}}\limits_{\pm 0.2}$ \\
\bottomrule
\end{tabular}%
}
\label{tab:imagenet100_10}%
\end{table}%
\vspace{-8mm}
\begin{table}[H]
\centering
\caption{Class incremental learning performance (top-5 / top-1 accuracy $_{\pm \text{std}}$\%) on ImageNet1000  with 10 incremental steps (100 classes per step). The best results are in bold.}
\resizebox{\textwidth}{29mm}{
\begin{tabular}{c|c|c|c|c|c|c|c|c|c|c|c}
\toprule
\#step & 1 & 2 & 3 & 4 & 5 & 6 & 7 & 8 & 9 & 10 & Avg \\
\midrule
LwF~\cite{li2017learning} & 90.1 &  77.7 & 63.9 & 51.8  & 43.0 & 35.5 & 31.6 & 28.4 & 26.4 & 24.3 & 42.5  \\
iCaRL~\cite{rebuffi2017icarl} & 90.0 & 83.0 & 77.5 & 70.5 & 63.0 & 57.5 & 53.5 & 50.0 & 48.0 & 44.0 & 60.8  \\
& -- & /57.9 & /48.8 & /40.9 & /35.5 & /31.8 & /28.8 & /25.5 & /24.2 & /22.7 & /35.1  \\
EEIL~\cite{castro2018end} & 94.9 & \textbf{94.9} & 84.7 & 77.8 & 71.7 & 66.8 & 62.5 & 59.0 & 55.2 & 52.3 & 69.4  \\
BiC~\cite{wu2019large} & 94.1 & 92.5 & 89.6 & \textbf{89.1} & 85.7 & 83.2 & 80.2 & 77.5 & 75.0 & 73.2 & 82.9  \\
IL2M~\cite{Belouadah_2019_ICCV} & -- & -- & -- & -- & -- & -- & -- & -- & -- & -- & 78.3  \\
& -- & /74.2 & /68.8 & /62.4 & /56.4 & /53.3 & /52.1 & /48.8 & /47.6 & /43.6 & /56.4  \\
WA~\cite{Zhao_2020_CVPR}  & 93.9 & 91.5 & 89.4 & 87.7 & 86.5 & 85.6 & 84.5 & 83.2 & 82.1 & 81.1 & 85.7 \\
& /79.8 & /75.3 & /70.9 & /68.1 & /65.6 & /63.6 & /61.2 & /59.2 & /57.4 & /55.6 & /64.1 \\
\midrule
EA (Ours)  & $\mathop{\text{94.4}}\limits_{\pm 0.3}$ & $\mathop{\text{92.5}}\limits_{\pm 0.1}$ & $\mathop{\textbf{90.4}}\limits_{\pm 0.1}$  & $\mathop{\text{89.0}}\limits_{\pm 0.1}$  & $\mathop{\textbf{87.7}}\limits_{\pm 0.0}$ & $\mathop{\textbf{86.8}}\limits_{\pm 0.1}$ &  $\mathop{\textbf{85.7}}\limits_{\pm 0.0}$ &  $\mathop{\textbf{84.5}}\limits_{\pm 0.0}$ &  $\mathop{\textbf{83.4}}\limits_{\pm 0.0}$  &  $\mathop{\textbf{82.6}}\limits_{\pm 0.1}$ & $\mathop{\textbf{87.0}}\limits_{\pm 0.0}$ \\
&  $\mathop{\text{/81.7}}\limits_{\pm 0.3}$ & $\mathop{\textbf{/76.6}}\limits_{\pm 0.5}$ & $\mathop{\textbf{/72.3}}\limits_{\pm 0.3}$  & $\mathop{\textbf{/69.7}}\limits_{\pm 0.2}$  & $\mathop{\textbf{/67.0}}\limits_{\pm 0.0}$  & $\mathop{\textbf{/65.0}}\limits_{\pm 0.1}$ & $\mathop{\textbf{/62.8}}\limits_{\pm 0.1}$ & $\mathop{\textbf{/61.0}}\limits_{\pm 0.1}$ & $\mathop{\textbf{/59.3}}\limits_{\pm 0.1}$  & $\mathop{\textbf{/57.8}}\limits_{\pm 0.1}$ & $\mathop{\textbf{/65.7}}\limits_{\pm 0.1}$ \\
\bottomrule
\end{tabular}%
}
\label{tab:imagenet1000_10}%
\end{table}%
\end{figure}

In this section, we make an effort to ameliorate the heavily biased models in CIL by the proposed energy aligning. We adopt the algorithm described in~\cite{wu2019large,Zhao_2020_CVPR} as baseline, which utilizes rehearsal~\cite{rebuffi2017icarl} and distillation ~\cite{hinton2015distilling,li2017learning} strategies. After training of each incremental step, we rectify the current model based on energy aligning. Specifically, the old and new categories in each step are regarded as two clusters ($M=2$), and the ``old cluster'' is viewed as the ``anchor cluster'' $\mathcal{O}_i$ described in Sec.~\ref{sec:es}. Then the shift scalar is calculated by Eq.~\eqref{eq:cluster_alpha}, and finally the model is corrected as Eq.~\eqref{eq:cluster_es}. The details and pseudo-code for CIL with energy aligning are presented in Appendix~\ref{app:td_il}.

\textbf{Datasets.} Experiments are conducted on the widely-used ImageNet ILSVRC 2012~\cite{Russakovsky2015} and CIFAR100~\cite{Krizhevsky09}. ImageNet ILSVRC 2012 includes about 1.2 million images for training and 50,000 images for validation in which two settings are provided --- ImageNet100 which contains 100 randomly selected classes and ImageNet1000 which consists of the whole 1,000 classes. CIFAR100 is comprised of 50,000 images for training and 10,000 images for evaluating with 100 classes.

\begin{figure}[t]
\begin{figure}[H]
\centering
\begin{minipage}[b]{0.7\textwidth}
\begin{figure}[H]
  \centering
  \subfloat[]{\label{fig:fre_main_a} \includegraphics[height=0.2\textwidth]{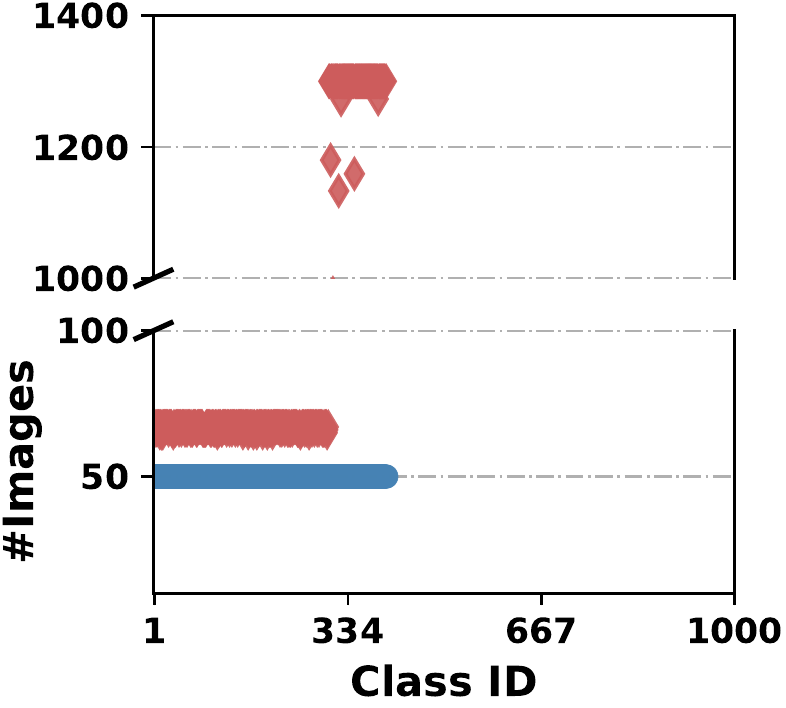}} \hspace{1mm}
  \subfloat[]{\label{fig:fre_main_b} \includegraphics[height=0.2\textwidth]{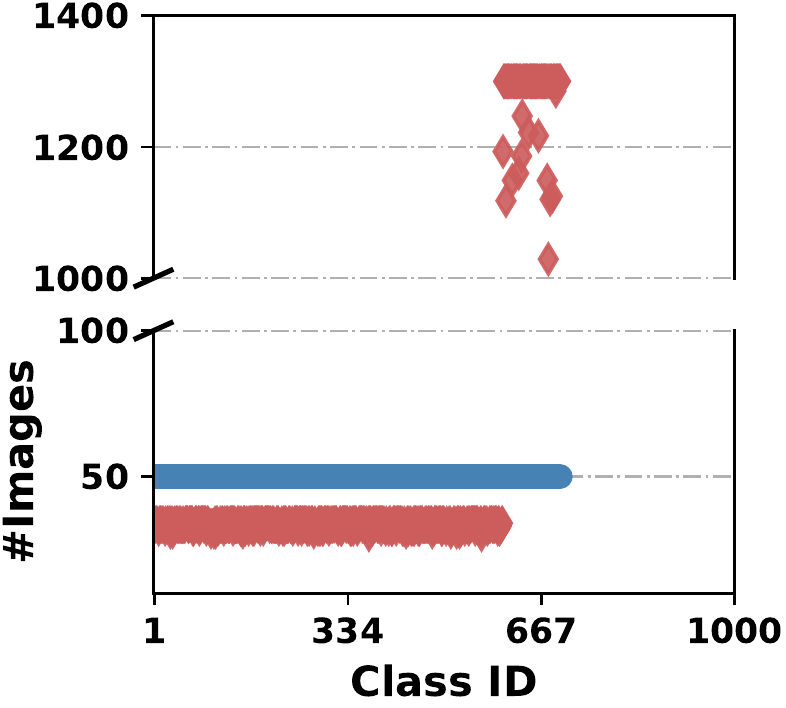}} \hspace{1mm}
  \subfloat[]{\label{fig:fre_main_c} \includegraphics[height=0.2\textwidth]{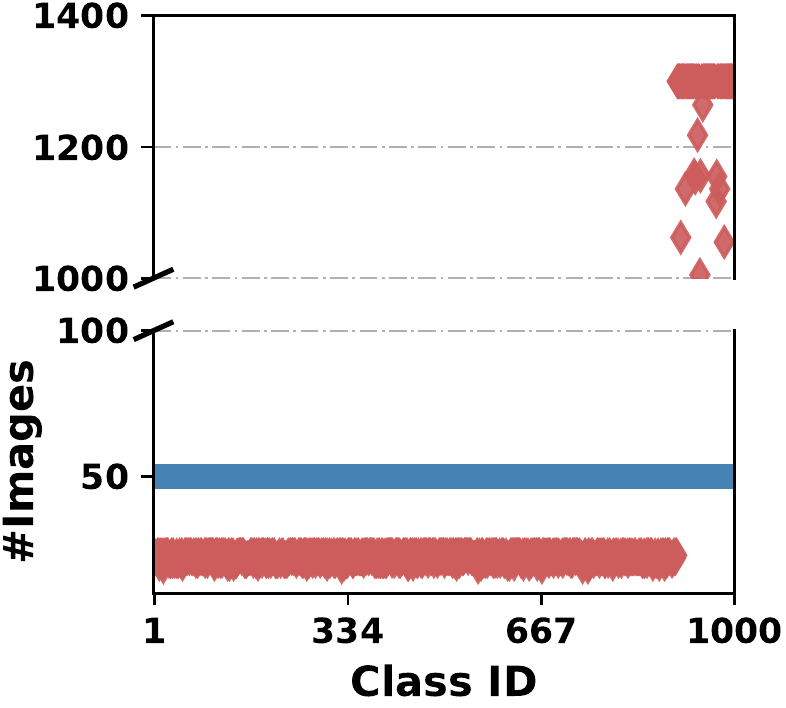}} \hspace{1mm}
  \subfloat[]{\label{fig:fre_main_d} \includegraphics[height=0.2\textwidth]{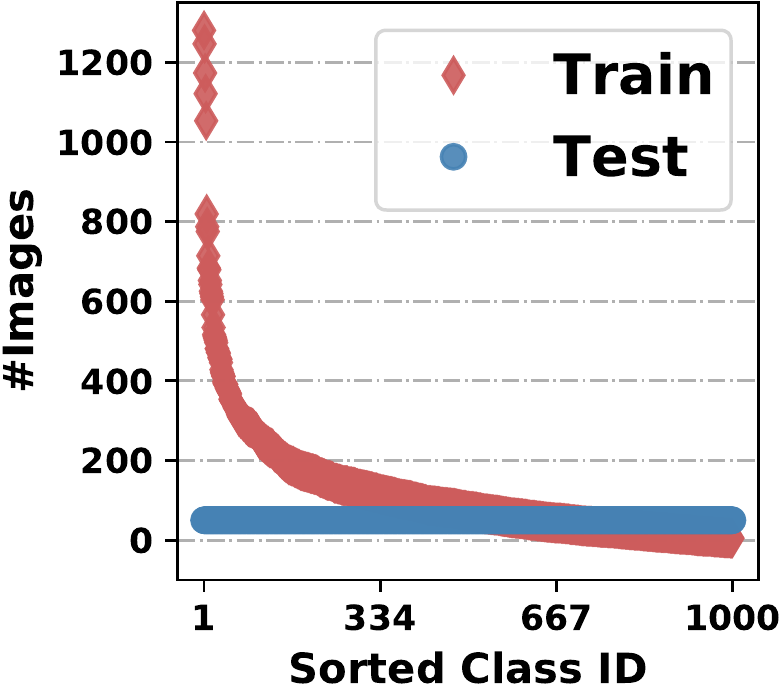}} \vspace{-2.5mm}
  \caption{Frequency distributions, (a), (b), (c) for the $4^{\text{th}}$, $7^{\text{th}}$, $10^{\text{th}}$ incremental step on ImageNet1000 (Sec.~\ref{sec:ex_il}); (d) ImageNet-LT (Sec.~\ref{sec:ex_lt}).}
\end{figure}
\vspace{-11mm}
\begin{figure}[H]
  \centering
  \subfloat[on ImageNet100]{
  \label{fig:il_eff_a} 
  \includegraphics[height=0.2\textwidth]{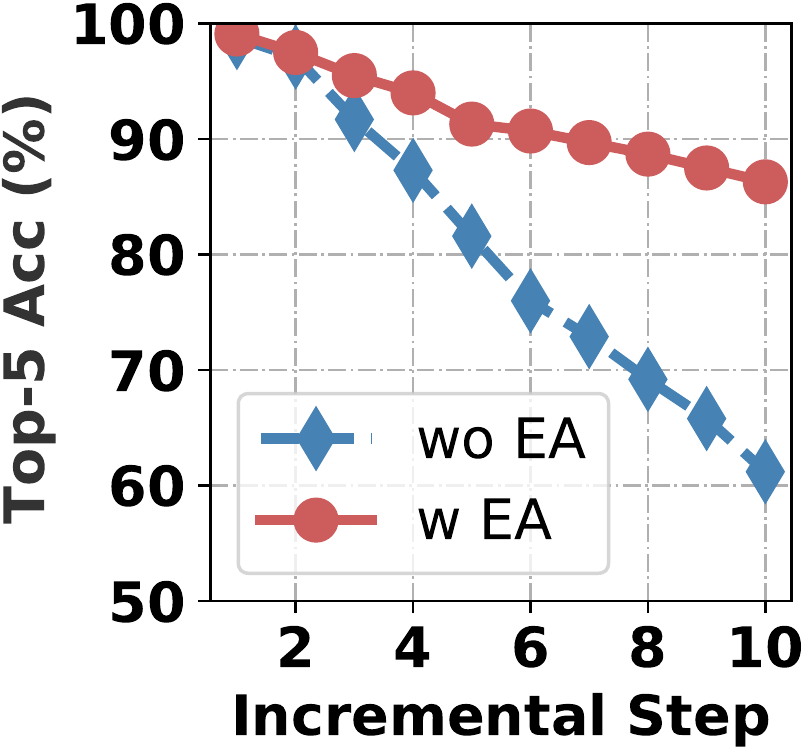} \hspace{1mm}
  \includegraphics[height=0.2\textwidth]{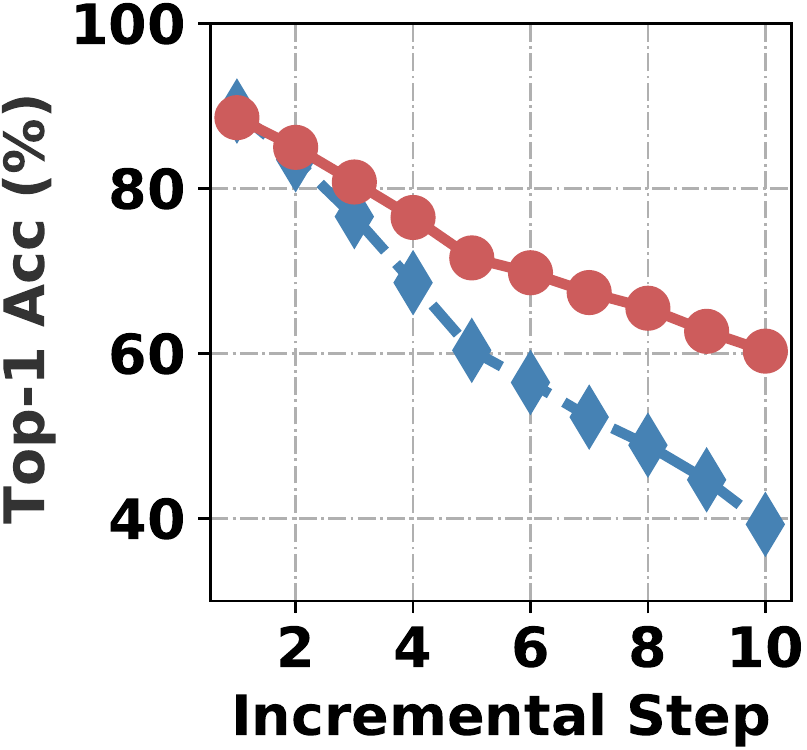}
  } \hspace{1mm}
  \subfloat[on ImageNet1000]{
  \label{fig:il_eff_b} 
  \includegraphics[height=0.2\textwidth]{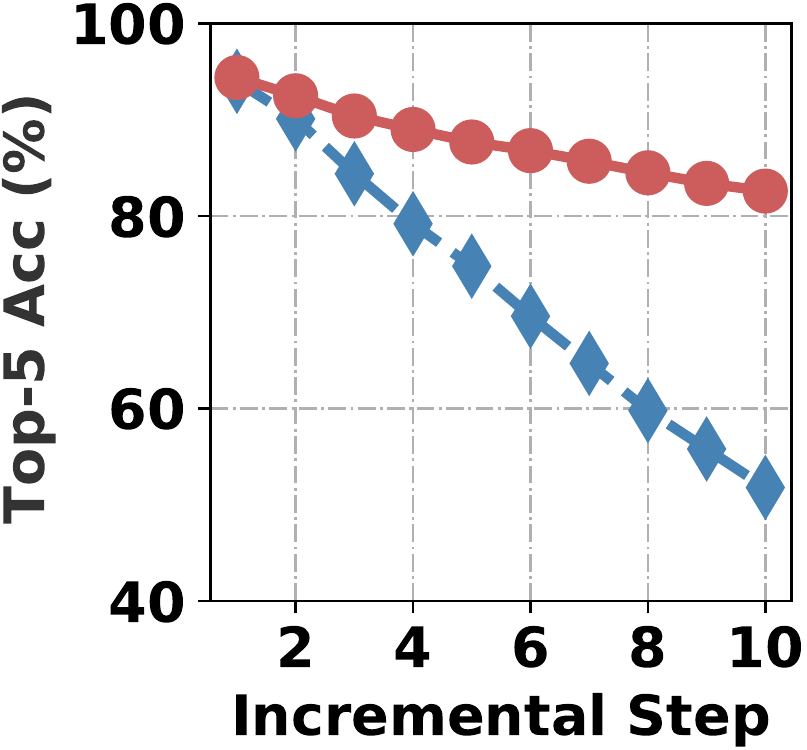} \hspace{1mm}
  \includegraphics[height=0.2\textwidth]{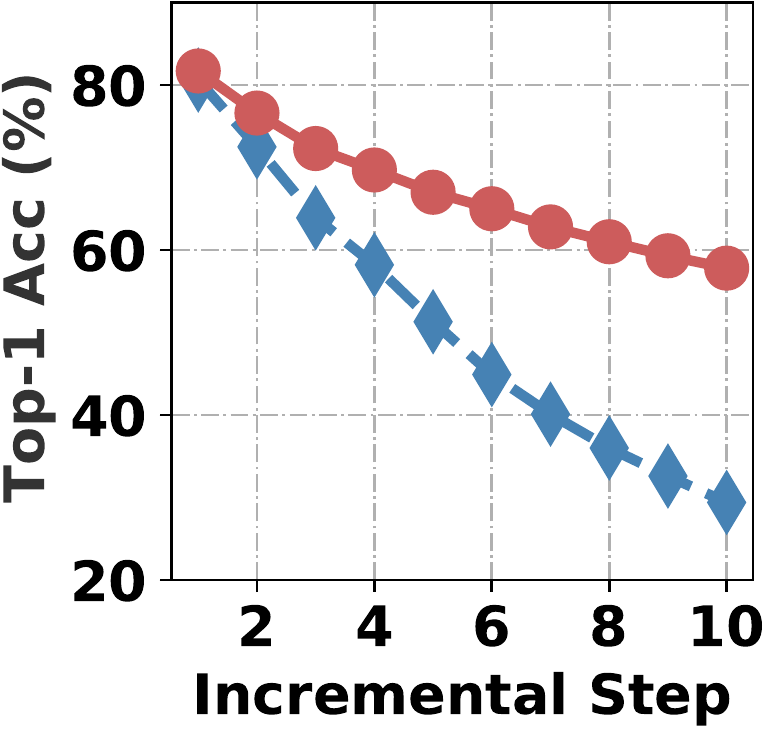}
  } \vspace{-2mm}
  \caption{Comparisons of learning with or without energy aligning.}
\end{figure}
\end{minipage}
\hspace{1mm}
\begin{minipage}[b]{0.28\textwidth}
  \centering
  \subfloat[without EA]{\label{fig:il_eff_c} \includegraphics[height=0.57\textwidth]{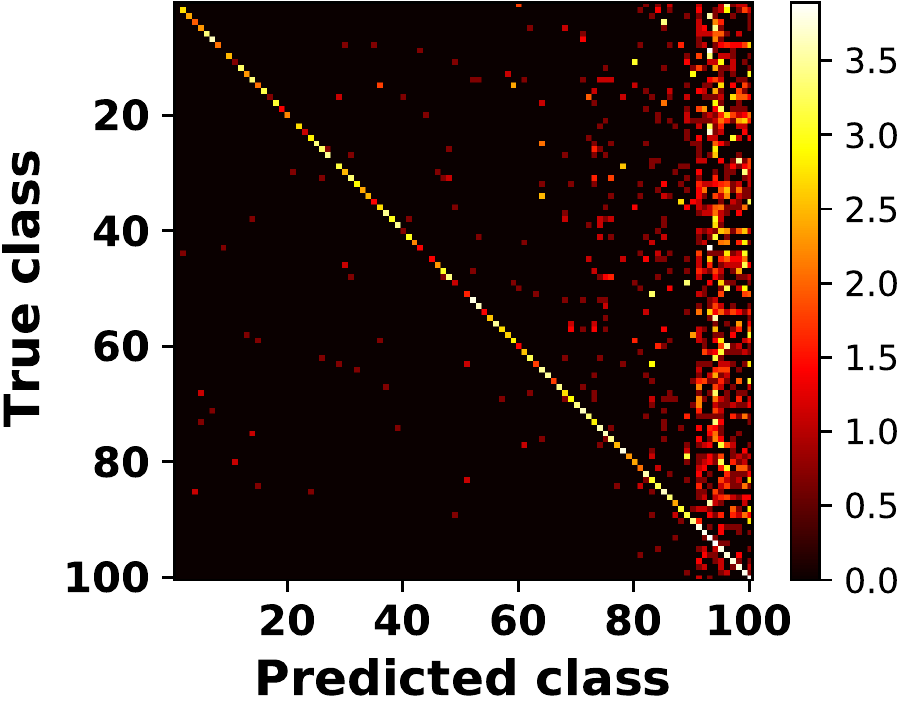}}\\ 
  \subfloat[with EA]{\label{fig:il_eff_d} \includegraphics[height=0.57\textwidth]{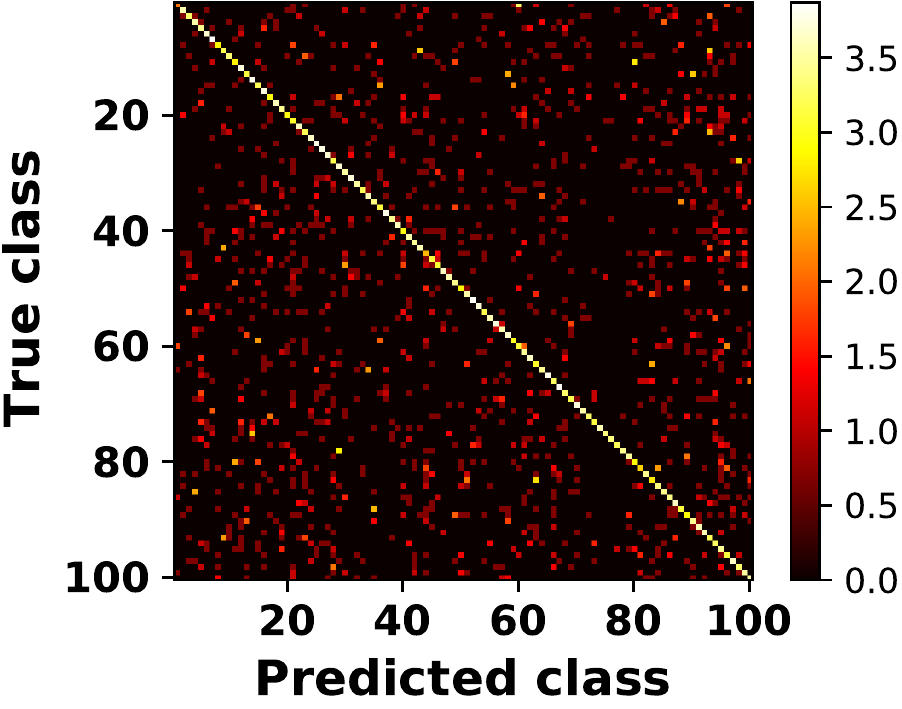}}\vspace{-2mm}
  \caption{Confusion matrix.}
\end{minipage}
\end{figure}
\vspace{-12mm}
\begin{table}[H]
\centering
\caption{Long-tailed recognition performance (top-1 accuracy $_{\pm \text{std}}$ \%) on iNaturalist. We present the results when training for 90 epochs / 200 epochs with ResNet-50. The best results are in bold.}
\begin{tabular}{l|cc|cc|cc|cc}
\toprule
Method &  \multicolumn{2}{c|}{Many}  &  \multicolumn{2}{c|}{Medium} &  \multicolumn{2}{c|}{Few}   &  \multicolumn{2}{c}{Overall} \\
\midrule
CE-DRW~\cite{cao2019learning} &    \multicolumn{2}{c|}{--}     &     \multicolumn{2}{c|}{--}    &    \multicolumn{2}{c|}{--}  & 63.7 & -- \\
CE-DRS~\cite{cao2019learning} &    \multicolumn{2}{c|}{--}     &    \multicolumn{2}{c|}{--}   &  \multicolumn{2}{c|}{--}    & 63.6 & --\\
CB-Focal~\cite{cui2019class} &  \multicolumn{2}{c|}{--}    & \multicolumn{2}{c|}{--}  &    \multicolumn{2}{c|}{--}   & 61.1 & -- \\
LDAM + DRW~\cite{cao2019learning} &   \multicolumn{2}{c|}{ --}    &   \multicolumn{2}{c|}{--}    &   \multicolumn{2}{c|}{--}   & 64.6 & / 66.1 \\
Decouple-NCM~\cite{DBLP:conf/iclr/KangXRYGFK20}   & 55.5 &/ 61.0 & 57.9 &/ 63.5 & 59.3 &/ 63.3 & 58.2 &/ 63.1 \\
Decouple-cRT~\cite{DBLP:conf/iclr/KangXRYGFK20}   & 69.0 &\textbf{/ 73.2} & 66.0 &/ 68.8 & 63.2 &/ 66.1 & 65.2 &/ 68.2 \\
Decouple-$\tau$-norm~\cite{DBLP:conf/iclr/KangXRYGFK20} & 65.6 &/ 71.1 & 65.3 &/ 68.9 & 65.9 &\textbf{/ 69.3} & 65.6 &/ 69.3 \\
Decouple-LWS~\cite{DBLP:conf/iclr/KangXRYGFK20}   & 65.0 &/ 71.0 & 66.3 &/ 69.8 & 65.5 &/ 68.8 & 65.9 &/ 69.5 \\
LA-loss~\cite{menon2020long}  &   \multicolumn{2}{c|}{--}    &    \multicolumn{2}{c|}{--}   &  \multicolumn{2}{c|}{--}  & 66.4 & -- \\
LA-loss + LDAM~\cite{menon2020long}  &   \multicolumn{2}{c|}{--}    &    \multicolumn{2}{c|}{--}   &  \multicolumn{2}{c|}{--}  & 68.4 & -- \\
LA-post-hoc~\cite{menon2020long}  &  66.1 & --    &  66.0 & --   & 66.0 & --  & 66.0 & -- \\
BBN~\cite{zhou2020bbn}  &   \multicolumn{2}{c|}{--}    &    \multicolumn{2}{c|}{--}   &  \multicolumn{2}{c|}{--}  & 66.3 &/ 69.6 \\
\midrule
EA (Ours)  &  $\mathop{\textbf{70.1}}\limits_{\pm 0.6}$ & $\mathop{\text{/ 73.0}}\limits_{\pm 0.1}$  &  $\mathop{\textbf{70.1}}\limits_{\pm 0.2}$ &$\mathop{\textbf{/ 71.0}}\limits_{\pm 0.1}$  & $\mathop{\textbf{69.5}}\limits_{\pm 0.3}$ &$\mathop{\text{/ 69.1}}\limits_{\pm 0.3}$  & $\mathop{\textbf{69.8}}\limits_{\pm 0.0}$ &$\mathop{\textbf{/ 70.4}}\limits_{\pm 0.1}$ \\
\bottomrule
\end{tabular}%
\label{tab:inat18}%
\end{table}
\end{figure}

\textbf{Protocols.} For fair comparisons, in accordance with the conventional experiment settings proposed in previous work~\cite{wu2019large,rebuffi2017icarl,castro2018end,Zhao_2020_CVPR},  ImageNet100 and ImageNet1000 are split into 10 incremental steps with 10 and 100 classes per step, respectively. In addition, 2,000 and 20,000 images are stored for old classes in the experiments on ImageNet100 and ImageNet1000, respectively. CIFAR100 is split into 2, 5, 10 and 20 steps, and 2,000 samples are stored in these experiments. We select rehearsal exemplars randomly. Some frequency distributions are illustrated in Fig.~\ref{fig:fre_main_a}, Fig.~\ref{fig:fre_main_b} and Fig.~\ref{fig:fre_main_c}, indicating the heavily imbalanced training data. More distributions are presented in Appendix~\ref{app:il_distri}. As new classes arrive, the number of samples can be retained for each class decreases gradually. Thus, the class imbalance problem becomes more serious, while the  target label distribution remains exact uniform for each class consistently. For each incremental step, the trained model is evaluated on all seen classes and reported on accuracy. After all, the average accuracy (Avg) over all incremental steps except the first step is calculated (as the first step is actually not related to ``incremental'').

\textbf{Implementation.} Our implementation is based on Pytorch~\cite{paszke2017automatic}. ResNet-18~\cite{He2015DeepRL,He2016IdentityMI} and ResNet-32 are employed as backbone for ImageNet and CIFAR100, respectively. More implementation details are provided in Appendix~\ref{app:td_il}.

\textbf{Effect of Energy Aligning.} We perform experiments on ImageNet100 and ImageNet1000 with 10 incremental steps respectively. As shown in Fig.~\ref{fig:il_eff_a} and Fig.~\ref{fig:il_eff_b}, EA significantly improves the performance (the gain in terms of top-5 accuracy in the last incremental step is more than 24\% on ImageNet100 and 29\% on ImageNet1000). We further plot the confusion matrices (after logarithmic transformation) for the models in the last incremental step on ImageNet100. From Fig.~\ref{fig:il_eff_c}, plain method (without EA) tends to predict objects as new classes, i.e., many samples from old classes (1\textasciitilde90) are misclassified as new classes (91\textasciitilde100). With the help of EA, the model treats new classes and old classes fairly as shown in Fig.~\ref{fig:il_eff_d}. These results intuitively show that EA can effectively alleviate class imbalance in CIL.

\textbf{Comparisons with State-of-the-Art.} Comparison results with the competitive and representative methods on ImageNet100 and Imagenet1000 are listed in Tab.~\ref{tab:imagenet100_10} and Tab.~\ref{tab:imagenet1000_10}, respectively, which apparently substantiate that our proposed algorithm achieves better performance when compared to state-of-the-art approaches, thereby demonstrating the effectiveness of EA again. Similarly, the proposed method also perform well under different settings on CIFAR100 (shown in Appendix~\ref{app:mr_il}).

\section{Application II: Long-Tailed Recognition}
\label{sec:ex_lt}

\begin{table}[t]
\centering
\caption{Long-tailed recognition performance (top-1 accuracy $_{\pm \text{std}}$ \%) on ImageNet-LT. We present the results of ResNet-50 / ResNeXt-50. The best results are in bold.}
\begin{tabular}{l|cc|cc|cc|cc}
\toprule
Methods & \multicolumn{2}{c|}{Many}  & \multicolumn{2}{c|}{Medium} & \multicolumn{2}{c|}{Few} & \multicolumn{2}{c}{Overall} \\
\midrule
Focal~\cite{lin2017focal} & -- &/ 64.3  & -- &/ 37.1  & -- &/ 8.2  & -- &/ 43.7  \\
OLTR~\cite{liu2019large}  & -- &/ 51.0   & -- &/ 40.8  & -- &/ 20.8  & -- &/ 41.9  \\
Decouple-OLTR~\cite{liu2019large,DBLP:conf/iclr/KangXRYGFK20} & 59.9 & --  & 45.8 & --  & 27.6 & --  & 48.7 & -- \\
Decouple-NCM~\cite{DBLP:conf/iclr/KangXRYGFK20} & 53.1 &/ 56.6  & 42.3 &/ 45.3  & 26.5 &/ 28.1 & 44.3 &/ 47.3 \\
Decouple-cRT~\cite{DBLP:conf/iclr/KangXRYGFK20} & 58.8 &/ 61.8 & 44.0 &/ 46.2 & 26.1 &/ 27.4 & 47.3 &/ 49.6 \\
Decouple-$\tau$-norm~\cite{DBLP:conf/iclr/KangXRYGFK20} & 56.6 &/ 59.1 & 44.2 &/ 46.9 & 27.4 &/ 30.7  & 46.7 &/ 49.4 \\
Decouple-LWS~\cite{DBLP:conf/iclr/KangXRYGFK20} & 57.1 &/ 60.2 & 45.2 &/ 47.2 & 29.3 &/ 30.3 & 47.7 &/ 49.9 \\
Capsule~\cite{sabour2017dynamic,liu2019large} & -- &/ 67.1  & -- &/ 40.0  & -- &/ 11.2  & -- &/ 46.5  \\
De-confound~\cite{tang2020long} & -- &\textbf{/ 67.9} & -- &/ 42.7 & -- &/ 14.7 & -- &/ 48.6 \\
Cosine-TDE~\cite{tang2020long} & -- &/ 61.8 & -- &/ 47.1 & -- &/ 30.4  & -- &/ 50.5 \\
Capsule-TDE~\cite{tang2020long} & -- &/ 62.3 & -- &/ 46.9 & -- &/ 30.6 & -- &/ 50.6 \\
De-confound-TDE~\cite{tang2020long} & -- &/ 62.7 & -- &/ 48.8 & -- &/ 31.6 & -- &/ 51.8 \\
LA-loss~\cite{menon2020long}  &   \multicolumn{2}{c|}{--}    &   \multicolumn{2}{c|}{--}  &  \multicolumn{2}{c|}{--}  & 51.1 & -- \\
LA-loss + LDAM~\cite{menon2020long}  &   \multicolumn{2}{c|}{--}    &    \multicolumn{2}{c|}{--}   &  \multicolumn{2}{c|}{--}  & 48.8 & -- \\
LA-post-hoc~\cite{menon2020long}  &  \textbf{62.3} & --    & 48.9 & --   &  30.2 & --  & 50.3 & -- \\
\midrule
EA (Ours)  &  $\mathop{\text{62.0}}\limits_{\pm 0.6}$ & $\mathop{\text{/ 62.1}}\limits_{\pm 0.6}$  &  $\mathop{\textbf{49.9}}\limits_{\pm 0.4}$ &$\mathop{\textbf{/ 50.2}}\limits_{\pm 0.2}$  & $\mathop{\textbf{34.6}}\limits_{\pm 1.1}$ &$\mathop{\textbf{/ 34.7}}\limits_{\pm 0.9}$  & $\mathop{\textbf{52.5}}\limits_{\pm 0.3}$ &$\mathop{\textbf{/ 52.7}}\limits_{\pm 0.1}$ \\
\bottomrule
\end{tabular}%
\label{tab:imagenet_lt}%
\end{table}%

Long-tailed distribution of training data is extremely challenging for DNNs, which is natural and common in real world. In this section, we rectify the biased models in long-tailed recognition by energy aligning. We first perform clustering based on the number of training samples per class. The cluster with the smallest average number of samples is treated as ``anchor cluster'' $\mathcal{O}_i$, then for each other cluster $\mathcal{O}_j$, the shift scalar can be obtained via Eq.~\eqref{eq:cluster_alpha}. Finally, the model can be corrected by Eq.~\eqref{eq:cluster_es} with the calculated shift scalars. The details and pseudo-code are presented in Appendix~\ref{app:td_lt}.

\textbf{Datasets.} Experiments are carried out on long-tailed datasets ImageNet-LT, iNaturalist 2018, CIFAR10-LT and CIFAR100-LT. ImageNet-LT~\cite{liu2019large}, artificially truncated from the original balanced ImageNet ILSVRC 2012~\cite{Russakovsky2015}, consists of 1000 classes with number of training images per class ranges from 1280 to 5 following a long-tailed distribution. The iNaturalist~\cite{van2018inaturalist} contains 8,142 classes, which is a real-world, naturally long-tailed dataset. CIFAR10-LT, CIFAR100-LT~\cite{cao2019learning} are the long-tailed versions of CIFAR10 and CIFAR100~\cite{Krizhevsky09} with specific degrees of data imbalance. The frequency distributions of ImageNet-LT is presented in Fig.~\ref{fig:fre_main_d}, and others are shown in Appendix~\ref{app:lt_distri}.

\textbf{Protocols.} After learning on the long-tailed training data, the models are evaluated on their corresponding balanced test (validation) data. In addition to the overall accuracy, the accuracies on three splits set based on class frequency --- Many-shot, Medium-shot and Few-shot are also reported.

\textbf{Implementation.} We implement energy aligning for long-tailed recognition with Pytorch~\cite{paszke2017automatic}. Following previous work~\cite{zhou2020bbn,tang2020long}, ResNet-50 and ResNeXt-50 (32x4d)~\cite{xie2017aggregated} are employed as backbones for ImageNet-LT, ResNet-50 for iNaturalist and ResNet-32 for CIFAR10-LT and CIFAR100-LT. All implementation details are provided in Appendix~\ref{app:td_lt}.

\begin{wrapfigure}[8]{r}{0.7\textwidth}
\vspace{-3mm}
\begin{minipage}[b]{0.43\textwidth}
    \centering
    \includegraphics[width=0.45\textwidth]{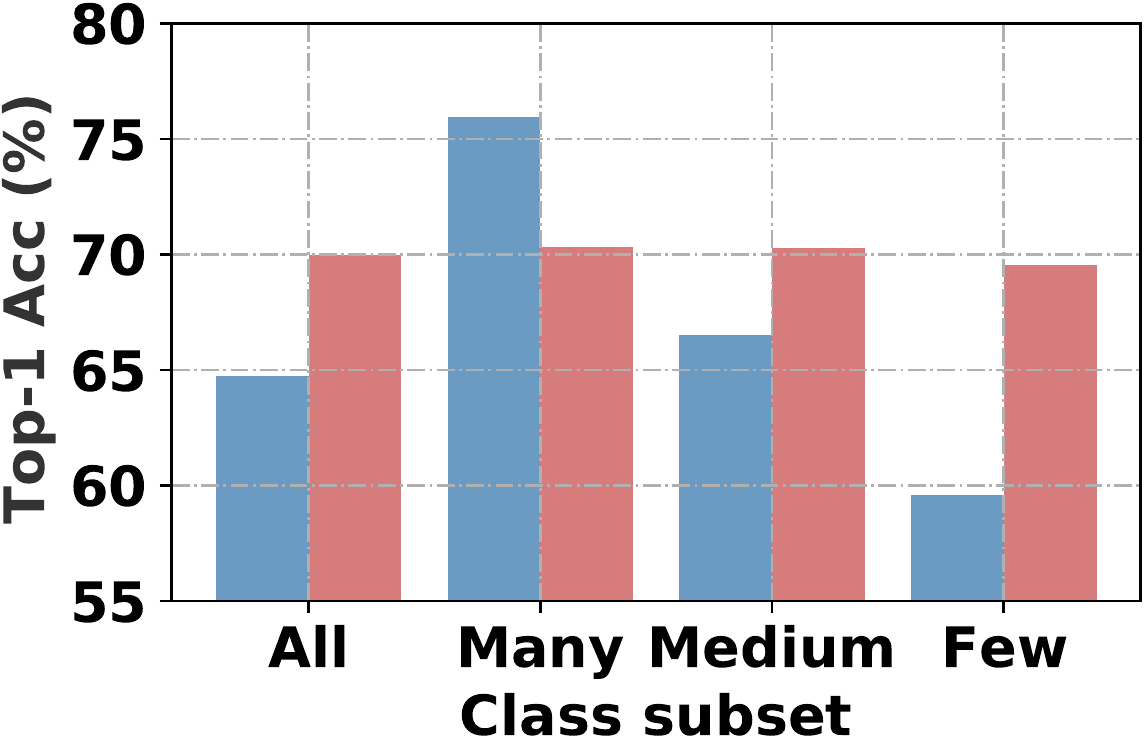} \hspace{0.5mm}
    \includegraphics[width=0.45\textwidth]{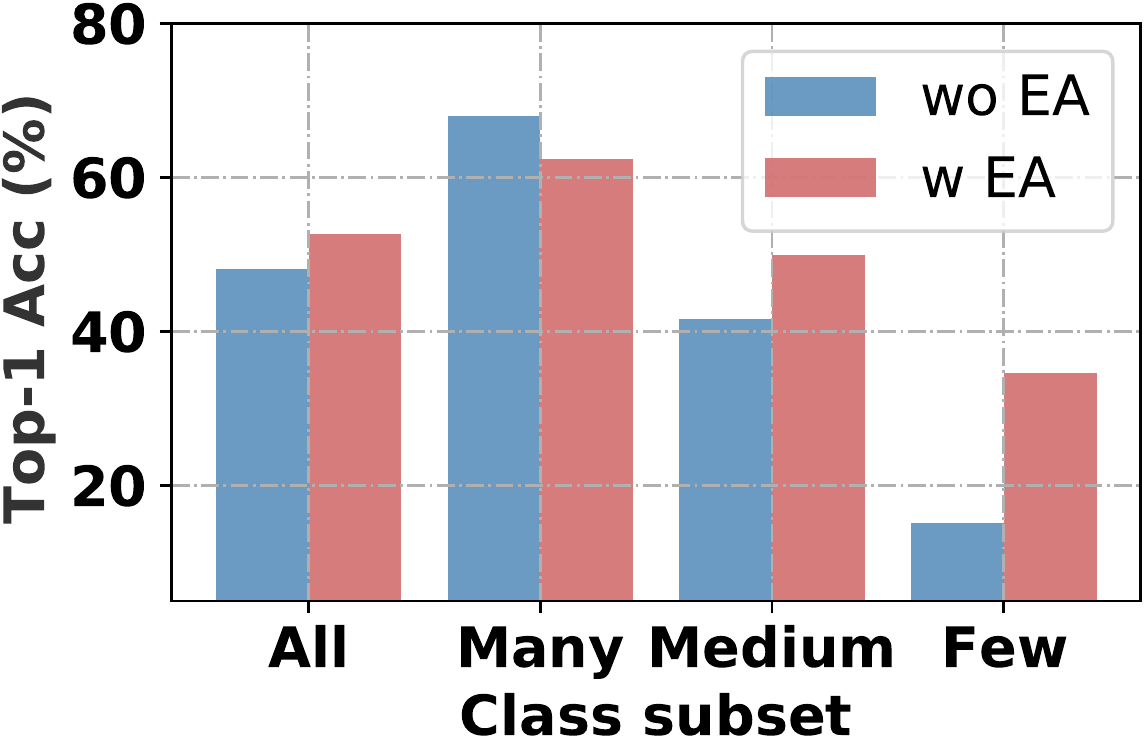} \vspace{-2mm}
    \caption{Learning with or without EA on iNaturalist (left) and ImageNet-LT (right).}
    \label{fig:lt_eff}
\end{minipage}
\hspace{1mm}
\begin{minipage}[b]{0.23\textwidth}
    \centering
    \includegraphics[width=0.83\textwidth]{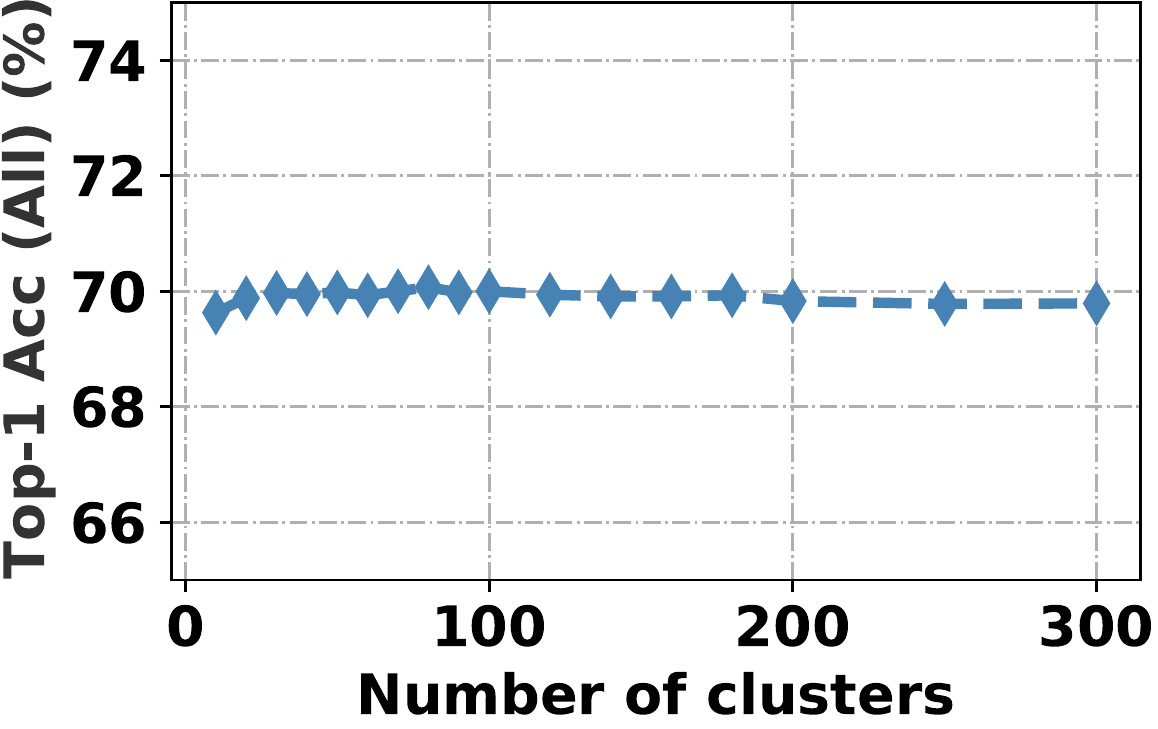} \vspace{-2mm}
    \caption{Influence of the number of clusters.}
    \label{fig:lt_num_clusters}
\end{minipage}
\end{wrapfigure}
\textbf{Effect of Energy Aligning.} The performance comparisons of models with or without energy aligning on iNaturalist (90 epochs, ResNet50) and ImageNet-LT (90 epochs, ResNeXt-50) are plotted in Fig.~\ref{fig:lt_eff}. With energy aligning, the model performs similarly on ``Many'', ``Medium'' and ``Few'' classes, and accomplishes better overall performance (5.25\% and 4.51\% gains in terms of top-1 accuracy on iNaturalist and ImageNet-LT, respectively), which demonstrates that the proposed energy aligning can effectively relieve the class imbalance problem in long-tailed recognition.

\textbf{Influence of number of clusters.} We investigate the influence of the number of clusters when performing energy aligning for long-tailed recognition. As depicted in Fig.~\ref{fig:lt_num_clusters}, the results on iNaturalist reveal that all choices for the number of clusters achieve similar performance, which indicates energy aligning in long-tailed recognition is insensitive to the number of clusters. In our experiments, we choose the number of clusters that make the model perform best on the sampled images $\{x_s\}_{s=1}^S$. 

More ablation studies are provided in Appendix, such as learning with different classifiers (Appendix~\ref{app:diff_cls}) and optimizers (Appendix~\ref{app:diff_optim}).

\textbf{Comparisons with State-of-the-Art.} We compare the performance of energy aligning to other recent studies that report state-of-the-art results on the commonly-used long-tailed benchmarks. The statistics are presented in Tab.~\ref{tab:inat18} for iNaturalist and Tab.~\ref{tab:imagenet_lt} for ImageNet-LT, respectively. For CIFAR10-LT and CIFAR100-LT, the results are provided in Appendix~\ref{app:mr_lt}. As evident from these tables, the proposed algorithm mostly surpasses the compared methods by a large margin.

\section{Related Work}

\textbf{Class Imbalance.} Class imbalance is naturally inherent in many tasks and applications. When learning on imbalanced dataset, the prediction results will erroneously tend to the majority classes while ignore the minority classes. The class imbalance problem is a classical and challenging topic, which has attracted extensive attention while it is still an open issue~\cite{zhang2019multi}. As described in Sec.~\ref{sec:intro}, over-sampling and under-sampling~\cite{shen2016relay,buda2018systematic,he2009learning,japkowicz2002class} are the most common strategies among data-level methods. As to algorithm-level approaches, apart from the prevalent cost-sensitive learning~\cite{huang2016learning,wang2017learning,byrd2019effect,ren2018learning,zhou2005training}, many recent studies incorporate transfer learning~\cite{yin2019feature,liu2019large}, metric learning~\cite{huang2016learning,oh2016deep,liu2019large}, meta learning~\cite{wang2017learning,liu2019large,shu2019meta}, hard negative mining~\cite{dong2017class,lin2017focal}, few-shot learning~\cite{liu2019large}, disentangled representation learning~\cite{DBLP:conf/iclr/KangXRYGFK20,zhou2020bbn} into the classical strategies to alleviate the severe bias.

\textbf{Class Incremental Learning.} In CIL, an ideal intelligent system is expected to be able to learn new knowledge continuously from the dynamic environment. However, the standard DNNs suffer from the notorious catastrophic forgetting problem~\cite{French1999CatastrophicFI,MichaelCatastrophic}, meaning that the DNNs tend to forget the old knowledge after learning the new ones. Recent methods alleviate catastrophic forgetting through parameter control~\cite{aljundi2018memory,kirkpatrick2017overcoming,zenke2017continual}, knowledge distillation~\cite{li2017learning,rebuffi2017icarl,zhou2019m2kd} and rehearsal~\cite{rebuffi2017icarl,wu2018incremental}. Another series of studies, like BiC~\cite{wu2019large}, IL2M~\cite{Belouadah_2019_ICCV}, WA~\cite{Zhao_2020_CVPR} and ScaIL~\cite{belouadah2020scail}, view class imbalance as an essential factor that causes catastrophic forgetting in class incremental learning, and put forward different solutions to address it.

\textbf{Long-Tailed Recognition.} In a long-tailed dataset, the number of training samples among different categories varies drastically. Long-tailed recognition with DNNs have attracted tremendous attention recently, and some long-tailed benchmarks have been established, such as long-tailed CIFAR10/100~\cite{cao2019learning}, long-tailed ImageNet~\cite{liu2019large} and iNaturalist~\cite{van2018inaturalist}. Latest studies~\cite{DBLP:conf/iclr/KangXRYGFK20,zhou2020bbn} disentangle the feature learning and the classifier training and achieve promising results. As to the understandings from other aspects, \cite{tang2020long} deals with this problem from the perspective of causal inference and \cite{menon2020long} presents a statistical framework to cope with the imbalanced label distribution of training data.

\textbf{Energy-Based Models.} EBMs supply a unified formulation for many probabilistic and non-probabilistic methods~\cite{lecun2006tutorial}. Recent studies have applied EBMs to several unsupervised learning tasks in computer vision, such as generative models~\cite{zhao2016energy,xie2018cooperative}, video generation~\cite{xie2019learning}, 3D shape synthesis and analysis~\cite{xie2018learning} and deep probabilistic regression~\cite{gustafsson2020energy,gustafsson2020train}. In~\cite{xie2016theory}, the authors show that a generative random field model can be derived from a discriminative model. Then, in~\cite{grathwohl2019your}, a standard discriminative model of $p_{\theta}(y | x)$ is reinterpreted as an energy-based model for the joint distribution $p_{\theta}(x, y)$, and a generative model is incorporated into optimization. In the recent work on out-of-distribution detection~\cite{liu2020energy}, the energy-based model is utilized to obtain the probability density $p_{\theta}(x)$ by marginalizing the joint distribution Eq.~\eqref{eq:joint_d} over $y$: $
p_{\theta}(x) = 
\int_{y' \in \mathcal{Y}} e^{- E_{\theta}(x, y')} /
Z_{\theta}
=
e^{- E_{\theta}(x)} /
Z_{\theta},
$ where, 
$
  E_{\theta}(x) = -
  \log \int_{y' \in \mathcal{Y}} e^{- E_{\theta}(x, y')}.
$
The above equation suggests that $-E_{\theta}(x)$ is linearly aligned with the log probability density $p_{\theta}(x)$ ideally. Thus, the free energy $E_{\theta}(x)$ is considered more suitable for out-of-distribution detection than the softmax score shown in Eq.~\eqref{eq:p_y_con_x}. The training objective is purely discriminative in~\cite{liu2020energy}. Inspired by it, starting by marginalizing the joint distribution Eq.~\eqref{eq:joint_d} over $x$, we derive an approach for correcting the biased models in this work.

\section{Conclusion and Further Work}
\label{sec:conclu_further}

In this paper, we propose a straightforward and effective algorithm to deal with the model bias issue when training on imbalanced dataset. From the perspective of energy-based models, we systematically analyze the relationship between the free energies of categories and the label distribution. Based on theoretical calculation, we propose the Energy Aligning (EA) approach to adjust output energies of different classes during inference to achieve better overall performance. EA is insensitive to hyper-parameters and by no means intervene training process, and the comprehensive experiments conducted on class incremental learning and long-tailed recognition benchmarks demonstrate that it outperforms many state-of-the-art methods. Though EA has achieved promising results, there are still many efforts need to be done, such as advanced training strategies of EBMs (like~\cite{grathwohl2019your}), more sophisticated methods to approximate the shift scalars. We hope that the work could encourage more studies to address class imbalance from the perspective of energy-based models.


\small{
\bibliographystyle{abbrv}
\bibliography{ref}
}







\newpage
\appendix

\section{Training Details for Learning with Energy Aligning}
\label{app:td}

\subsection{Details for Class Incremental Learning}
\label{app:td_il}

We present the learning process of class incremental learning with energy aligning in Algorithm~\ref{algo:es_il}. Assume that $B$ batches of training data $\{\mathcal{D}^{(b)}\}_{b=1}^{B}$ from different categories are available gradually in which, for the $b^{\text{th}}$ incremental step, the new training data $\mathcal{D}^{(b)}$ comes from $C^{(b)}$ new classes, and the rehearsal data $\mathcal{D}^{(b)}_{old}$ (which is selected from the previous data $\{\mathcal{D}^{(1)}, \cdots, \mathcal{D}^{(b-1)}\}$) comes from $C^{(b)}_{old}$ classes, where $C^{(1)}_{old}=0$ and $C^{(b)}_{old} = \sum_{k=1}^{b-1}C^{(k)}, b>1$. As described in Sec.~\ref{sec:ex_il}, the number of rehearsal samples in each incremental step (except the first step, as there is no old data in the initial step) is constant, i.e., $|\mathcal{D}_{old}^{(1)}|=0$, $|\mathcal{D}_{old}^{(2)}| = |\mathcal{D}_{old}^{(3)}| = \cdots = |\mathcal{D}_{old}^{(B)}|$, so that as more categories are encountered, the number of rehearsal samples per old class decreases, and the problem of class imbalance becomes more serious. 

The model $f_{\theta}(\cdot)$ is trained with a compound loss:
\begin{equation}
  \mathcal{L}_{CIL}(x, y^*) = (1 - \lambda) \mathcal{L}_{CIL-CE}(x, y^*) + \lambda \mathcal{L}_{CIL-KD}(x),
\label{eq:cil_loss}
\end{equation}
where $y^*$ denotes the true label, $(x, y^*) \in \mathcal{D}^{(b)} \cup \mathcal{D}^{(b)}_{old}$,\ $\lambda$ is used to balance the two losses which is set to $\lambda_{base} \cdot \frac{C^{(b)}_{old}}{C^{(b)}+C^{(b)}_{old}}$ with a hyper-parameter $\lambda_{base}$. The cross-entropy loss is defined as
\begin{equation}
  \mathcal{L}_{CIL-CE}(x, y^*) = \sum_{i=1}^{C^{(b)}+C^{(b)}_{old}} - \mathbb{I}_{i=y^*} \log( p_{\theta}(i|x) ),
\label{eq:ce_loss}
\end{equation}
where $(x, y^*) \in \mathcal{D}^{(b)} \cup \mathcal{D}^{(b)}_{old}$, \ $\mathbb{I}_{i=y^*}$ is the indicator function and $p_{\theta}(i|x)$ is the predicted probability of the $i^{\text{th}}$ class defined as Eq.~\eqref{eq:p_y_con_x}. The knowledge distillation loss is defined as
\begin{equation}
  \mathcal{L}_{CIL-KD}(x) = \sum_{i=1}^{C^{(b)}_{old}} - \hat{q}_{\theta_t;\alpha}(i|x) \log ( q_{\theta}(i|x) ),
\label{eq:kd_loss}
\end{equation}

\begin{equation*}
\hat{q}_{\theta_t;\alpha}(i|x) =
\frac{
e^{ f_{\theta_{t};\alpha}(x)[i] / T}
}{
\sum_{j=1}^{C^{(b)}_{old}}  e^{ f_{\theta_{t};\alpha}(x)[j] / T}
}
, \quad
q_{\theta}(i|x) =
\frac{
e^{f_{\theta}(x)[i] / T}
}{
\sum_{j=1}^{C^{(b)}_{old}}  e^{f_{\theta}(x)[j] / T}
},
\end{equation*}
where $x \in \mathcal{D}^{(b)} \cup \mathcal{D}^{(b)}_{old}$, \ $T$ represents the temperature of knowledge distillation. The teacher model $f_{\theta_{t};\alpha}(x)$ is obtained from the $(b-1)^{\text{th}}$ incremental step, which is corrected by a shift scalar $\alpha$ (in class incremental learning,  the old classes form the ``old cluster'' which is viewed as ``anchor cluster'' (described in Sec.~\ref{sec:es}), and the new classes form the ``new cluster'', then, a shift scalar $\alpha$ can be calculated for the ``new cluster'').

\begin{algorithm}[htb]
\centering
\caption{Class Incremental Learning with Energy Aligning}
\begin{algorithmic}[1]
\State \textbf{Input:} Training data $\{ \mathcal{D}^{(b)}\}_{b=1}^{B}$ of $B$ incremental steps
\State \textbf{Initialization:} Model $f_{\theta}$
\State $f_{\theta} \leftarrow Train(f_{\theta}, \mathcal{D}^{(1)})$ \algorithmiccomment{{\color{gray}Train with loss Eq.~\eqref{eq:ce_loss} (no rehearsal data in the first step)}}
\State $f_{\theta_t;\alpha} \leftarrow DetachCopy(f_{\theta})$ \algorithmiccomment{{\color{gray}Save as the teacher model for the next incremental step;\\ \hspace{14mm} a dummy shift scalar $\alpha$ is added here for the convenience of the following statement}}
\State $\mathcal{D}_{re}^{(2)} \leftarrow RandomSample(\mathcal{D}^{(1)})$ \algorithmiccomment{{\color{gray}Select rehearsal samples for the next incremental step}}
\For{$b \in 2, \cdots, B $}
\algorithmiccomment{{\color{gray}For each incremental step}}
  \State $f_{\theta} \leftarrow Train(f_{\theta}, f_{\theta_t;\alpha}, \mathcal{D}^{(b)}, \mathcal{D}_{re}^{(b)})
  $ \algorithmiccomment{{\color{gray}Train with loss Eq.~\eqref{eq:cil_loss}}}
  \State $\mathcal{D}_{ea} \leftarrow RandomSample(\mathcal{D}^{(b)}, \mathcal{D}_{re}^{(b)})$ \algorithmiccomment{{\color{gray}Select samples for energy aligning}}
  \State $\alpha \leftarrow ShiftScalar(f_{\theta}, \mathcal{D}_{ea})$  \algorithmiccomment{{\color{gray}Calculate shift scalar by Eq.~\eqref{eq:cluster_alpha}}}
  \State $f_{\theta_t;\alpha} \leftarrow DetachCopy(Correct(f_{\theta}, \alpha))$ \algorithmiccomment{{\color{gray}Save the model corrected by Eq.~\eqref{eq:cluster_es} as \\ \hspace{84mm} teacher model used in the next step}}
  \State $\mathcal{D}_{re}^{(b+1)} \leftarrow RandomSample(\mathcal{D}^{(b)}, \mathcal{D}_{re}^{(b)})$ \algorithmiccomment{{\color{gray}Select rehearsal samples from current \\ \hspace{94mm} new data and rehearsal data}}
\EndFor
\State \textbf{Output:} Model $f_{\theta_t;\alpha}$
\end{algorithmic}
\label{algo:es_il}
\end{algorithm}

Unless otherwise specified, our implementation is based on Pytorch~\cite{paszke2017automatic}. SGD is used to train models with momentum 0.9 and the cosine classifier is employed. Random crop and horizontal flip are utilized to augment training images. Since the model will be trained for several incremental steps, the base weight decay $r_{base}$ is reduced to $r^{(b)}$ in the $b^{\text{th}}$ incremental step by $r^{(b)} = r_{base} \cdot \eta^{b-1}$. For ImageNet, we adopt ResNet-18~\cite{He2015DeepRL,He2016IdentityMI} as backbone with batch size set to 256, learning rate starting from 0.1 and attenuating by 1/10 successively after 30, 60, 80 and 90 epochs (100 epochs in total),  $\lambda_{base}$ set to 1.0, $r_{base}$ set to 0.0005, $\eta$ set to 0.5 for ImageNet100 and 0.25 for ImageNet1000, respectively. For CIFAR100, a 32-layer ResNet is employed as backbone with batch size set to 32, learning rate starting from 0.1 and attenuating by 1/10 successively after 100, 150 and 200 epochs (250 epochs in total); for 2 steps, $\lambda_{base}=1.0, r_{base}=0.0001, \eta=0.3$; for 5 steps, $\lambda_{base}=0.75, r_{base}=0.0002, \eta=0.3$; for 10 and 20 steps, $\lambda_{base}=0.5, r_{base}=0.0004, \eta=0.5$. All models are trained on a single Tesla V100 GPU. $T$ is set to 2.0 in all experiments. The augmentations employed on the sampled data for energy aligning are listed in Tab.~\ref{tab:ea_aug} (all the used transformations for augmentation are implemented in torchvision, see in \url{https://pytorch.org/vision/stable/transforms.html}). It is worth noting that all the experiments on different datasets and settings for class incremental learning and long-tailed recognition are based on the same augmentation strategies, which shows that the proposed energy aligning is robust. 

All the datasets used in our experiments for class incremental learning are available for free to researchers for non-commercial use, and do not contain personally identifiable information or offensive content.

\subsection{Details for Long-Tailed Recognition}
\label{app:td_lt}
  
The detailed learning process of long-tailed recognition with energy aligning is described in Algorithm~\ref{algo:es_lt}. Assume that training data $\mathcal{D}$ including $C$ categories are prepared, the model $f_{\theta}(\cdot)$ is trained with the standard cross-entropy loss:
\begin{equation}
  \mathcal{L}_{LT-CE}(x, y^*) = \sum_{i=1}^{C} - \mathbb{I}_{i=y^*} \log( p_{\theta}(i|x) ),
\label{eq:lt_ce_loss}
\end{equation}
where $\mathbb{I}_{i=y^*}$ is the indicator function and $p_{\theta}(i|x)$ is the predicted probability of the $i^{\text{th}}$ class defined as Eq.~\eqref{eq:p_y_con_x}.

Unless otherwise specified, our implementation is based on Pytorch~\cite{paszke2017automatic}. SGD is used to train models with momentum 0.9 and the cosine classifier is employed. Random crop and horizontal flip are utilized to augment training images. Weight decay is set to 0.0005. For ImageNet-LT, ResNet-50 and ResNeXt-50 (32x4d)~\cite{xie2017aggregated} are adopted as backbones which are trained for 90 epochs with batch size set to 512. We use the cosine learning rate schedule gradually decaying from 0.2 to 0. For iNaturalist, ResNet-50 is employed as backbone which are trained for 90 epochs and 200 epochs, with batch size set to 512. We use the cosine learning rate schedule gradually decaying from 0.1 to 0. For CIFAR10-LT and CIFAR100-LT, ResNet-32 used in~\cite{zhou2020bbn} are employed as the backbone which are trained for 200 epochs with batch size of 128. The initial learning rate is set to 0.1 and the first five epochs is trained with the linear warm-up learning rate schedule, and then the learning rate is decayed at the $120^{\text{th}}$ and $160^{\text{th}}$ epoch by 0.01 following~\cite{zhou2020bbn,tang2020long}. We use the Jenks natural breaks algorithm\footnote{\url{https://pypi.org/project/jenkspy/}}~\cite{jenks1967data} to divide the categories into $M$ clusters based on the number of training samples per class. The models for CIFAR10-LT and CIFAR100-LT are trained on a single Tesla V100 GPU, while for ImageNet-LT and iNaturalist are trained on two Tesla V100 GPUs. 

All the datasets used in our experiments for long-tailed recognition are available for free to researchers for non-commercial use, and do not contain personally identifiable information or offensive content.

\begin{algorithm}[htb]
\centering
\caption{Long-Tailed Recognition with Energy Aligning}
\begin{algorithmic}[1]
\State \textbf{Input:} Training data $\mathcal{D}$
\State \textbf{Initialization:} Model $f_{\theta}$
\State $f_{\theta} \leftarrow Train(f_{\theta}, \mathcal{D})$ \algorithmiccomment{{\color{gray}Train with loss Eq.~\eqref{eq:lt_ce_loss}}}
\State $\mathcal{D}_{ea} \leftarrow RandomSample(\mathcal{D})$ \algorithmiccomment{{\color{gray}Select samples for energy aligning}}
\State $\alpha_1, \alpha_2, \cdots, \alpha_{M-1}  \leftarrow ShiftScalar(f_{\theta}, \mathcal{D}_{ea})$  \algorithmiccomment{{\color{gray}Calculate shift scalars by Eq.~\eqref{eq:cluster_alpha}}}
\State $f_{\theta;\{\alpha_j\}_{j=1}^{M-1}} \leftarrow Correct(f_{\theta}, \alpha_1, \alpha_2, \cdots, \alpha_{M-1})$ \algorithmiccomment{{\color{gray}Correct the biased model by Eq.~\eqref{eq:cluster_es}}}
\State \textbf{Output:} Model $f_{\theta;\{\alpha_j\}_{j=1}^{M-1}}$
\end{algorithmic}
\label{algo:es_lt}
\end{algorithm}

\begin{table}[t]
  \centering
  \caption{List of transformations used for estimating shift scalars.}
  \begin{tabular}{c p{4.5cm} c p{1.8cm}}
  \toprule
       Transformation & Description & Parameter & Range  \\
  \midrule
      Flip-x  & Flip the image horizontally \\
      Flip-y  & Flip the image vertically \\
      Rotate & Rotate the image by $R$ degrees & $R$ & [-10, 10] \\
      Affine & Affine transformation of the image by $R$ degrees & $R$ & [-10, 10] \\
      Perspective & Perspective transformation of the image with the degree $R$ of distortion.& $R$ & [0, 0.5] \\
      Brightness & Adjust the brightness of the image with factor $R$  & $R$  & [0, 5] \\
      Contrast  & Control the contrast of the image with factor $R$  & $R$  & [0, 5] \\
      Saturation  & Adjust the saturation of the image with factor $R$  & $R$  & [0, 5] \\
      Hue  & Control the hue of the image with factor $R$  & $R$  & [0, 0.5] \\
      Erase & Select a rectangle region with scale $R_1$ and ratio $R_2$ in the image and erase its pixels. & $R_1, R_2$ & [0.02, 0.33], [0.3, 3.3] \\
  \bottomrule
  \end{tabular}
  \label{tab:ea_aug}
\end{table}

\section{Frequency Distributions of the Datasets.}
\label{app:distri}

\subsection{Frequency Distributions for Class Incremental Learning}
\label{app:il_distri}

\begin{figure}[t]
\centering
\begin{minipage}[b]{0.99\textwidth}
\centering
\includegraphics[height=0.2\textwidth]{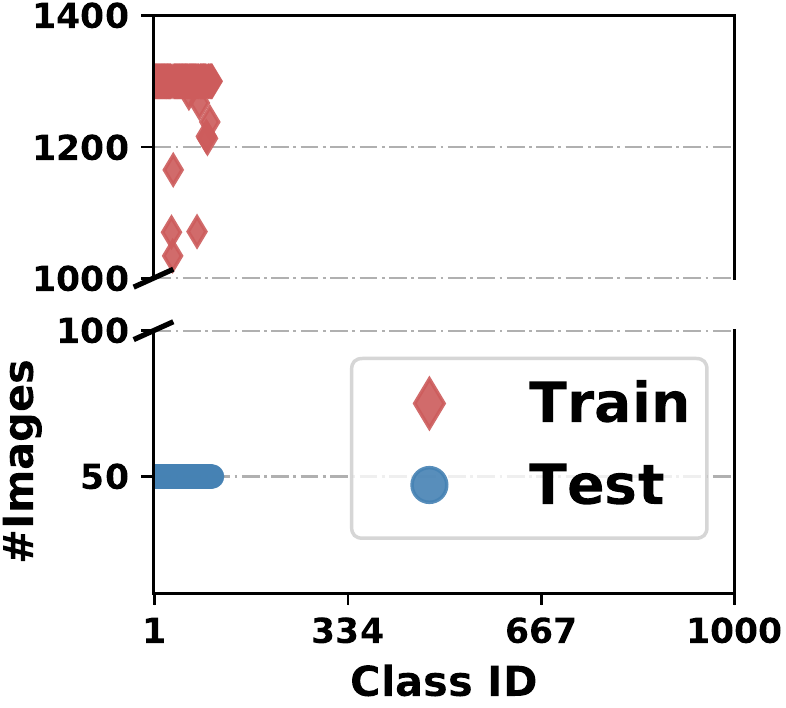}
\hspace{0.5mm}
\includegraphics[height=0.2\textwidth]{./fig/freil_imagenet_1000_10_3}
\hspace{0.5mm}
\includegraphics[height=0.2\textwidth]{./fig/freil_imagenet_1000_10_6}
\hspace{0.5mm}
\includegraphics[height=0.2\textwidth]{./fig/freil_imagenet_1000_10_9}
\caption{Frequency distributions in the $1^{\text{th}}$, $4^{\text{th}}$, $7^{\text{th}}$ and $10^{\text{th}}$ incremental step (from left to right) on ImageNet1000 with 10 incremental steps in total (100 classes per step).}
\label{fig:il_distri_imagenet1000}
\end{minipage}
\vspace{0.2mm}\\
\begin{minipage}[b]{0.99\textwidth}
\centering
\includegraphics[height=0.2\textwidth]{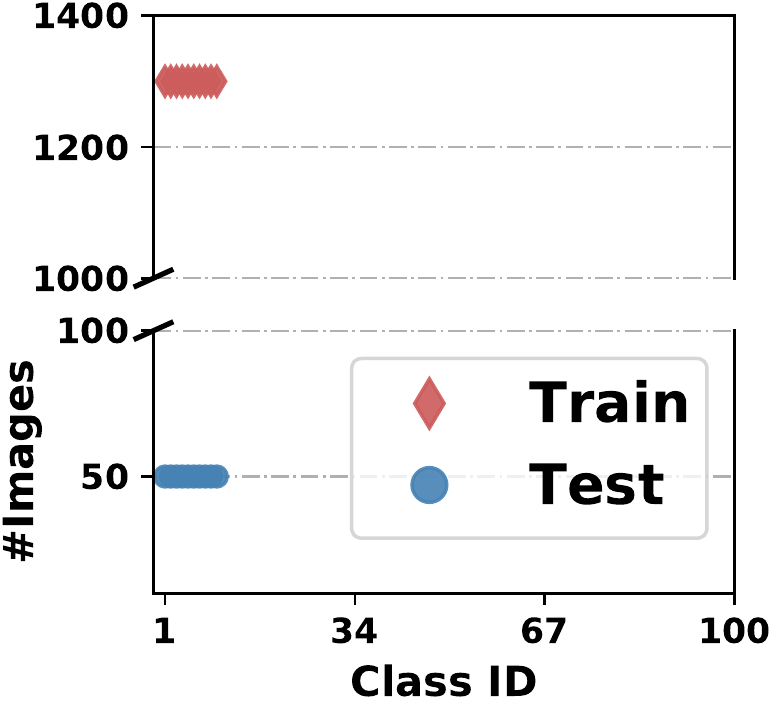}
\hspace{0.5mm}
\includegraphics[height=0.2\textwidth]{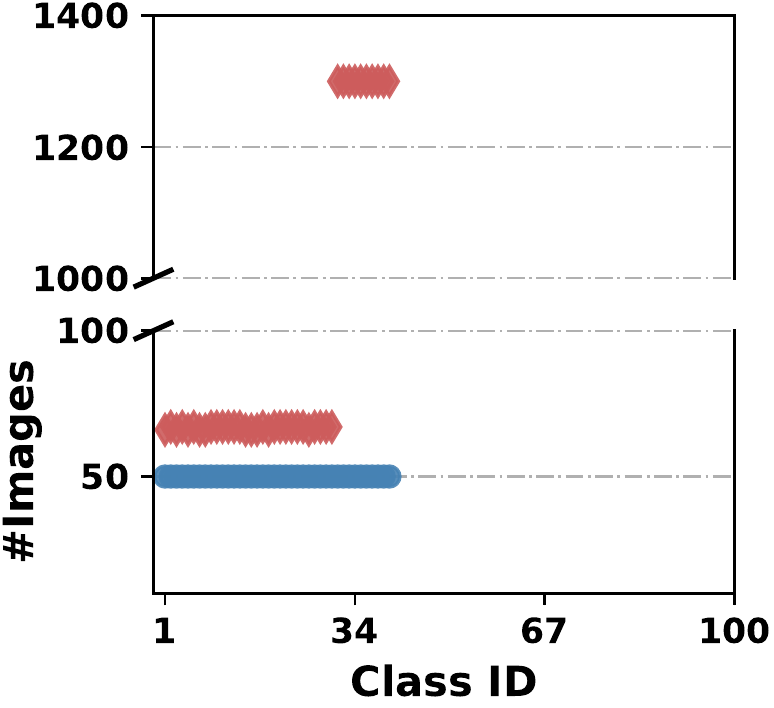}
\hspace{0.5mm}
\includegraphics[height=0.2\textwidth]{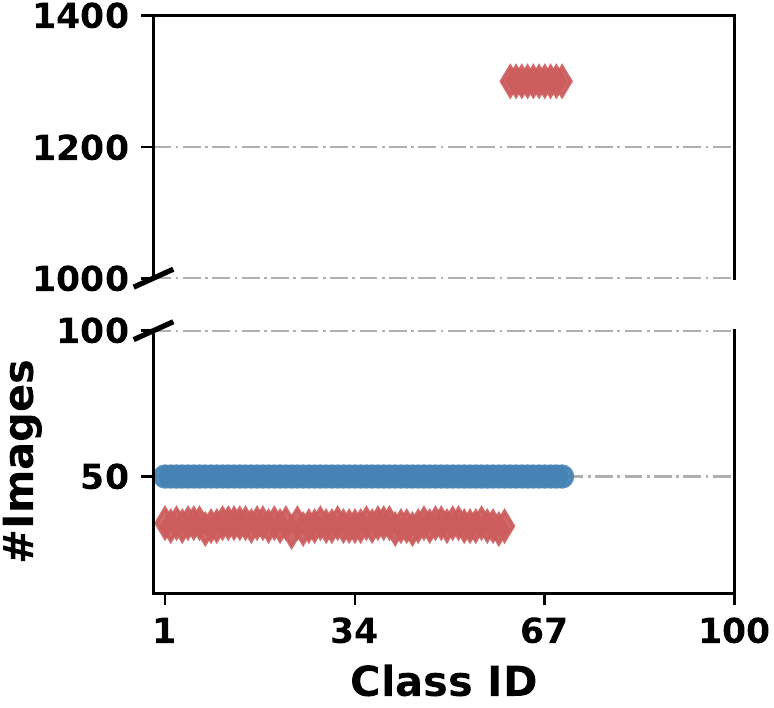}
\hspace{0.5mm}
\includegraphics[height=0.2\textwidth]{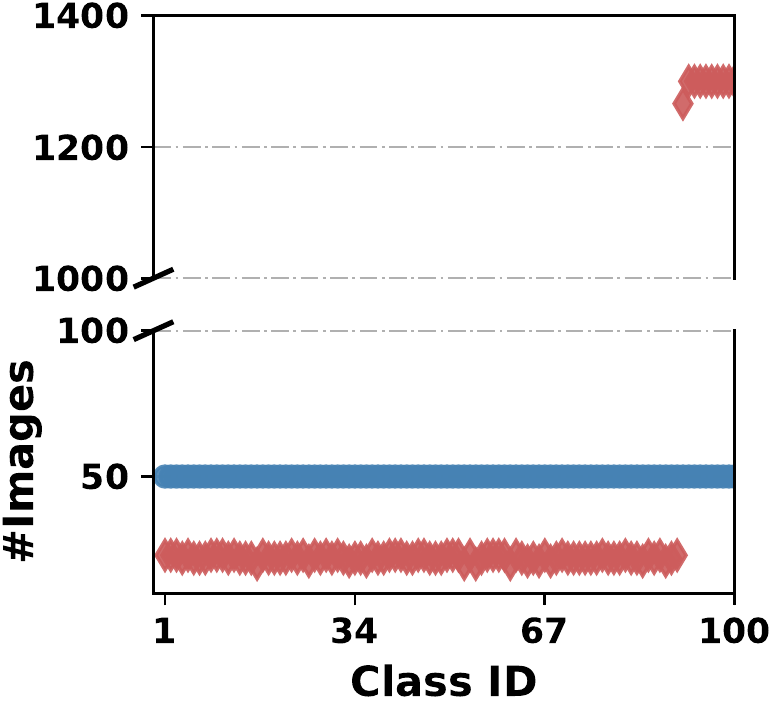}
\caption{Frequency distributions in the $1^{\text{th}}$, $4^{\text{th}}$, $7^{\text{th}}$ and $10^{\text{th}}$ incremental step (from left to right) on ImageNet100 with 10 incremental steps in total (10 classes per step).}
\label{fig:il_distri_imagenet100}
\end{minipage}
\vspace{0.2mm}
\begin{minipage}[b]{0.99\textwidth}
\centering
\subfloat[In the $1^{\text{th}}$, $7^{\text{th}}$, $13^{\text{th}}$ and $20^{\text{th}}$ incremental step (from left to right, 20 incremental steps in total with 5 classes per step).
]{
\includegraphics[height=0.2\textwidth]{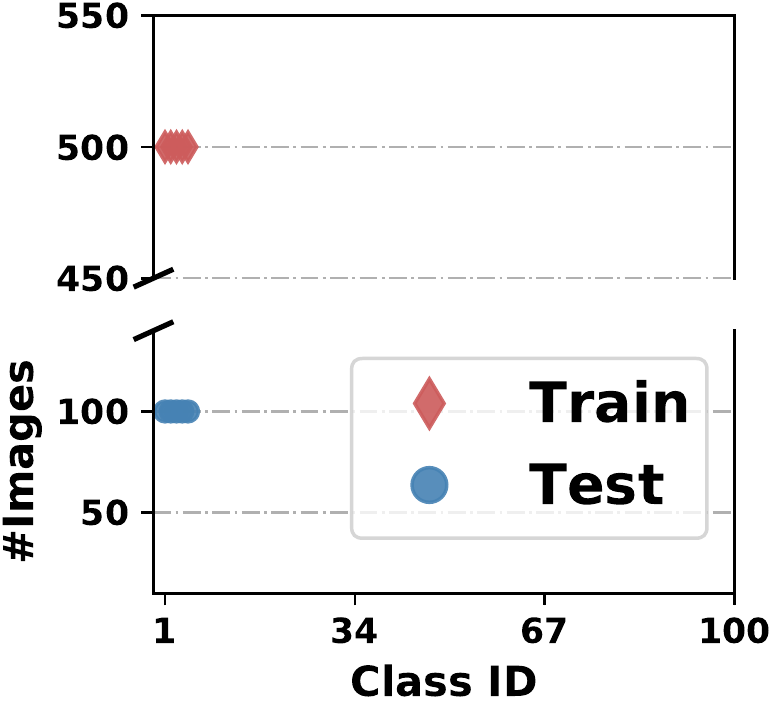}
\hspace{0.5mm}
\includegraphics[height=0.2\textwidth]{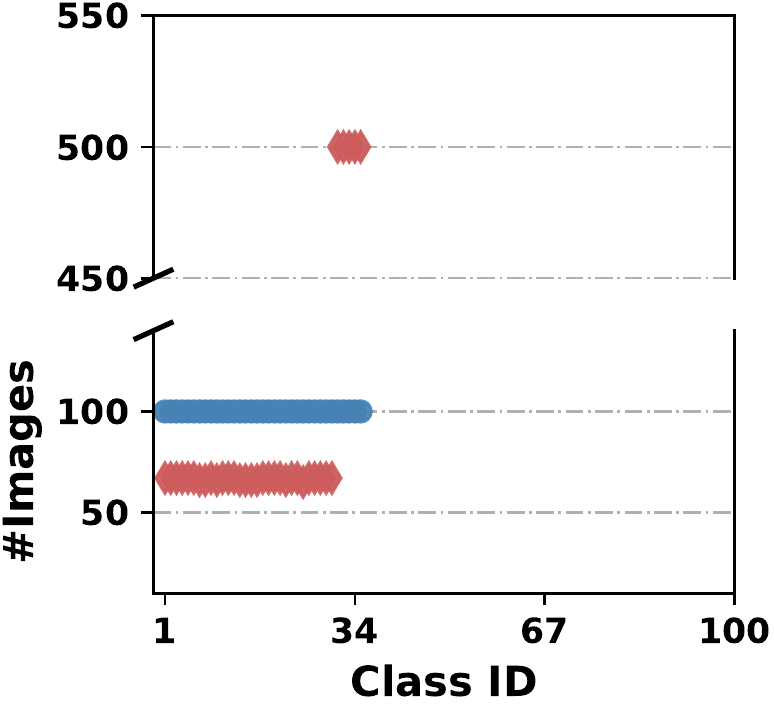}
\hspace{0.5mm}
\includegraphics[height=0.2\textwidth]{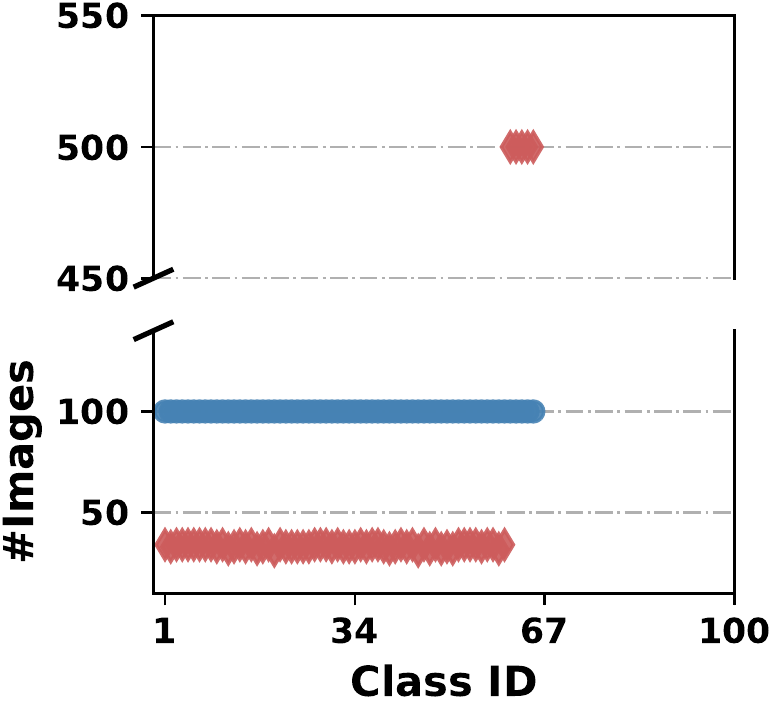}
\hspace{0.5mm}
\includegraphics[height=0.2\textwidth]{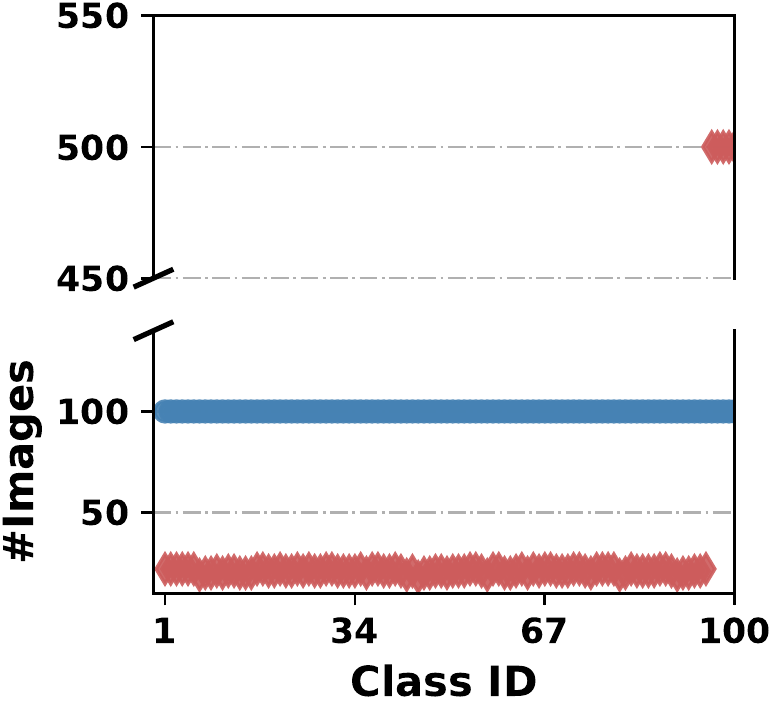}
}
\\
\subfloat[In the $1^{\text{th}}$, $4^{\text{th}}$, $7^{\text{th}}$ and $10^{\text{th}}$ incremental step (from left to right, 10 incremental steps in total with 10 classes per step).
]{
\includegraphics[height=0.2\textwidth]{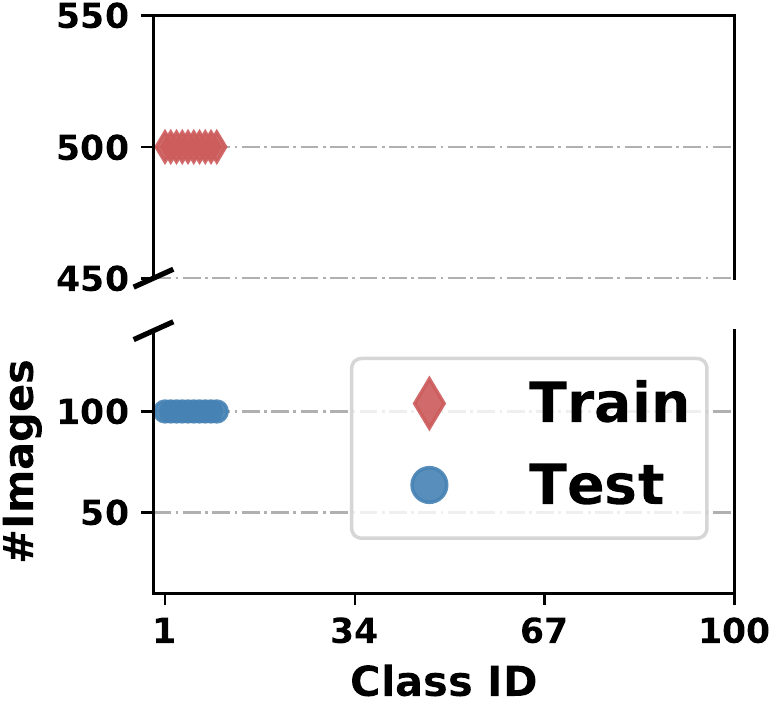}
\hspace{0.5mm}
\includegraphics[height=0.2\textwidth]{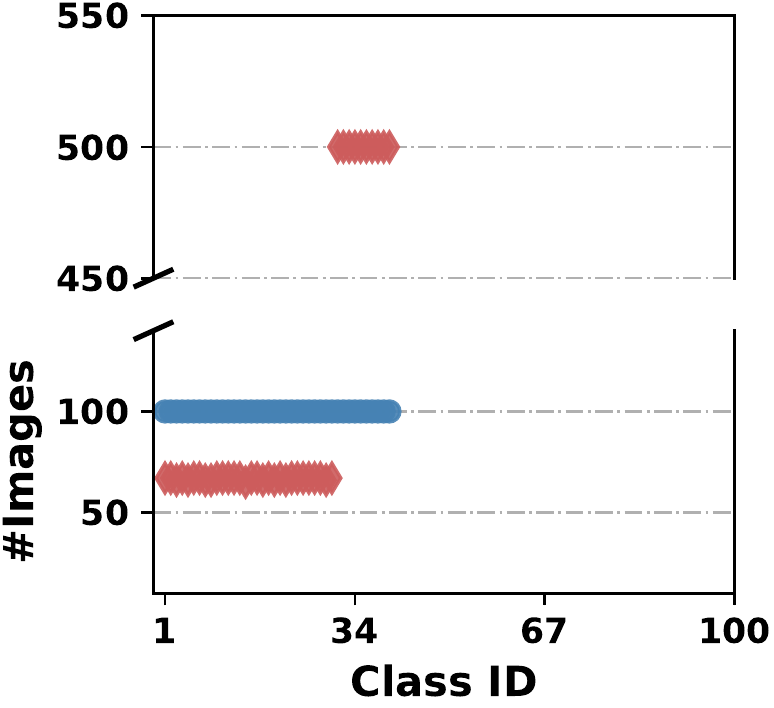}
\hspace{0.5mm}
\includegraphics[height=0.2\textwidth]{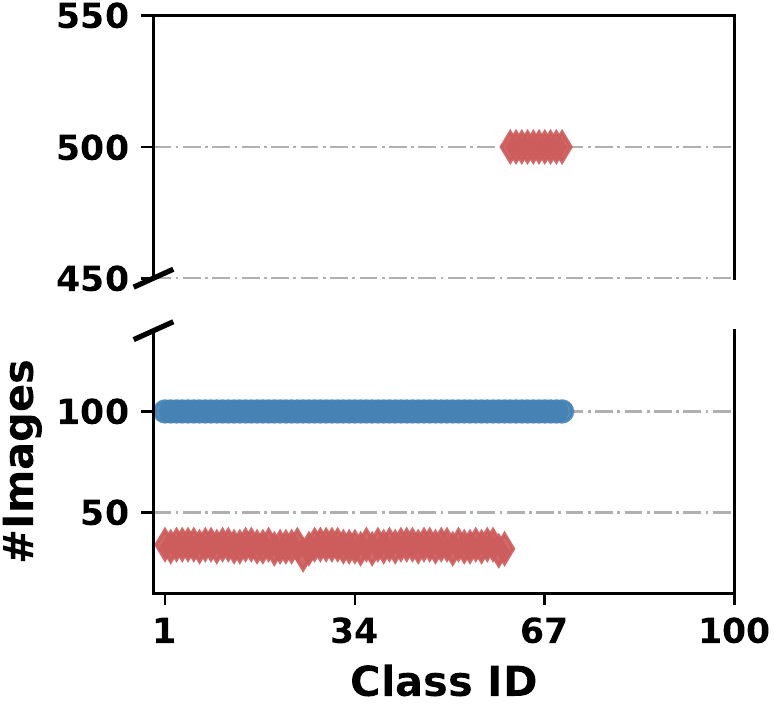}
\hspace{0.5mm}
\includegraphics[height=0.2\textwidth]{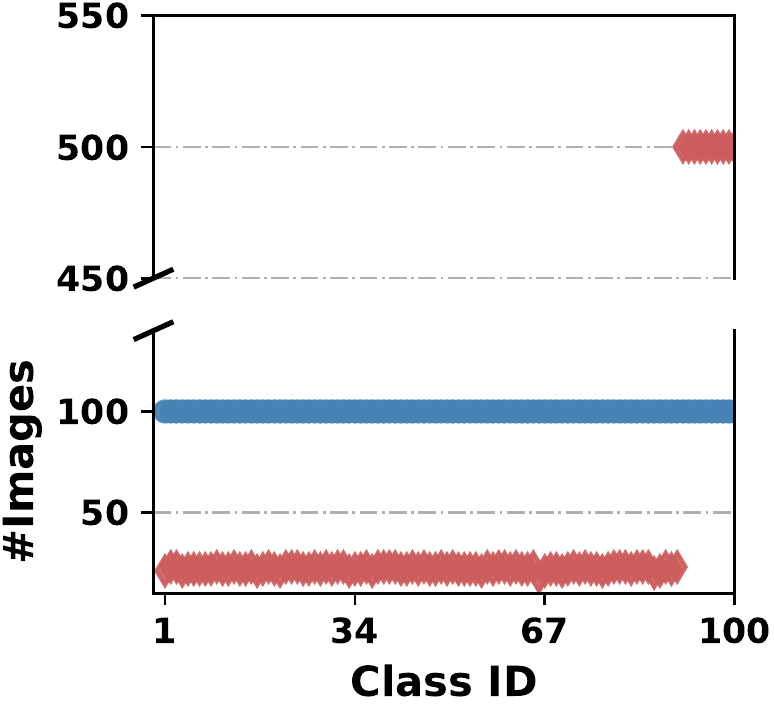}
}
\\
\subfloat[In the $1^{\text{th}}$ (left) and $5^{\text{th}}$ (right) incremental step (5 incremental steps in total with 20 classes per step).
]{
\includegraphics[height=0.2\textwidth]{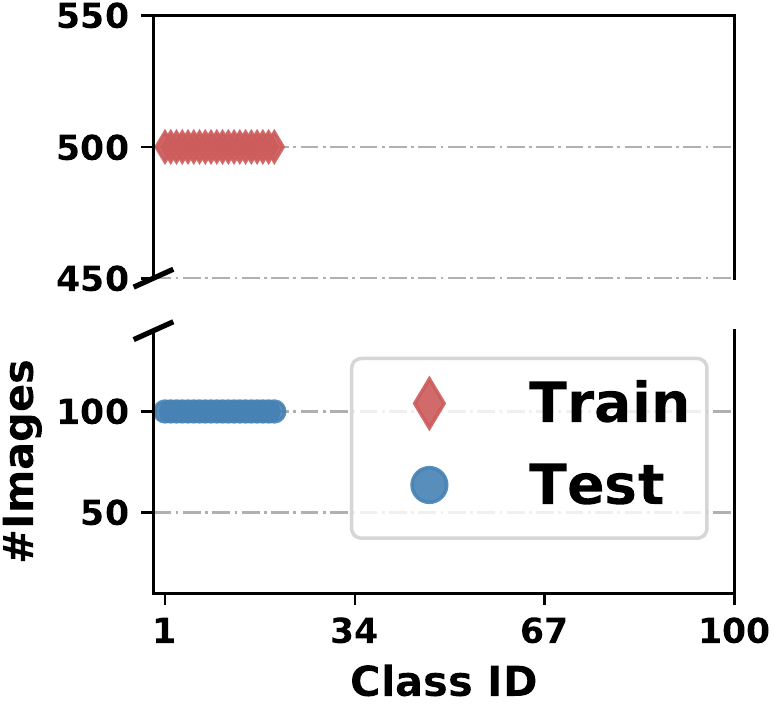}
\includegraphics[height=0.2\textwidth]{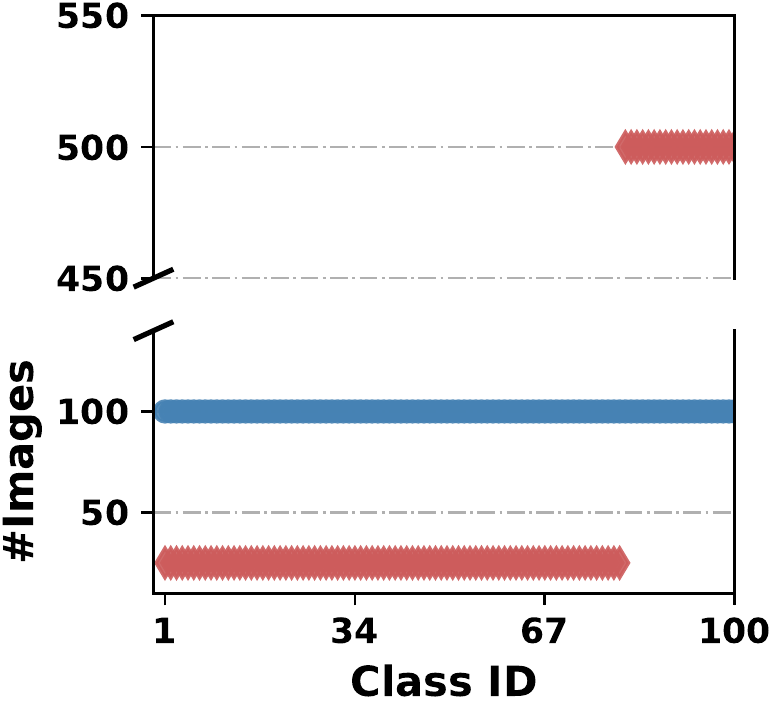}
}
\hspace{2mm}
\subfloat[In the $1^{\text{th}}$ (left) and $2^{\text{th}}$ (right) incremental step (2 incremental steps in total with 50 classes per step).
]{
\includegraphics[height=0.2\textwidth]{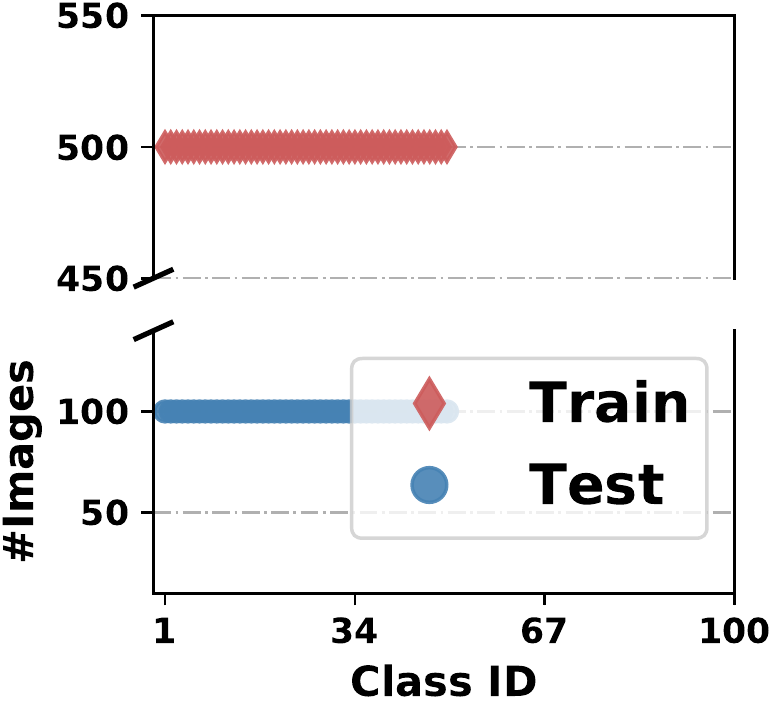}
\includegraphics[height=0.2\textwidth]{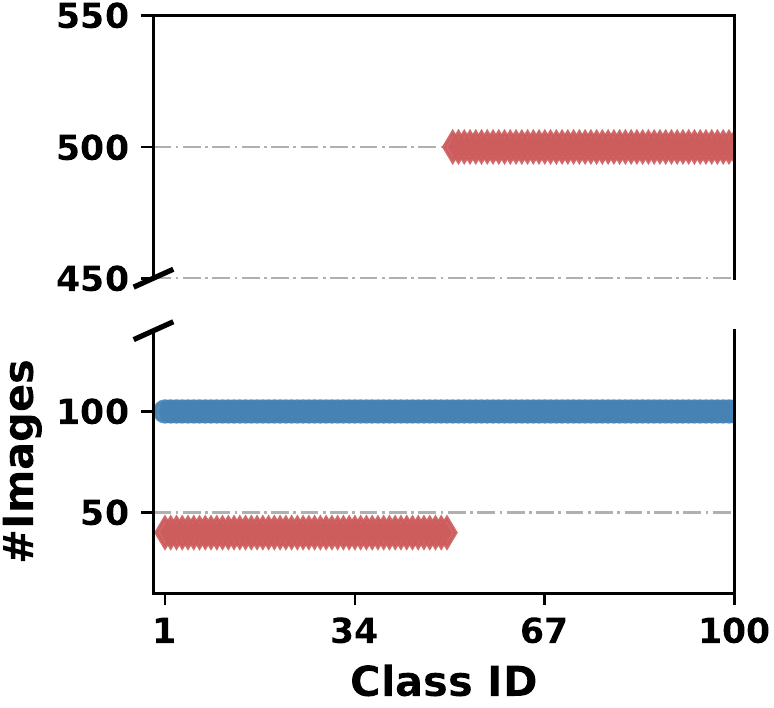}
}
\caption{Frequency distributions in different incremental steps on CIFAR100.}
\label{fig:il_distri_cifar}
\end{minipage}
\end{figure}

For class incremental learning, we plot the frequency distributions of ImageNet1000 with 10 incremental steps in Fig.~\ref{fig:il_distri_imagenet1000}, ImageNet100 with 10 incremental steps in Fig.~\ref{fig:il_distri_imagenet100}, and  CIFAR100 with 20, 10, 5 and 2 incremental steps in Fig.~\ref{fig:il_distri_cifar}. It can be seen that the problem of class imbalance becomes more serious as more classes are involved.

\subsection{Frequency Distributions for Long-Tailed Recognition}
\label{app:lt_distri}

\begin{figure}[t]
\centering
\begin{minipage}[b]{0.27\textwidth}
\centering
\includegraphics[height=0.78\textwidth]{./fig/fre_imagenet}
\caption{Frequency distributions of ImageNet-LT.}
\label{fig:lt_distri_imagenet}
\end{minipage}
\hspace{1mm}
\begin{minipage}[b]{0.7\textwidth}
\includegraphics[height=0.295\textwidth]{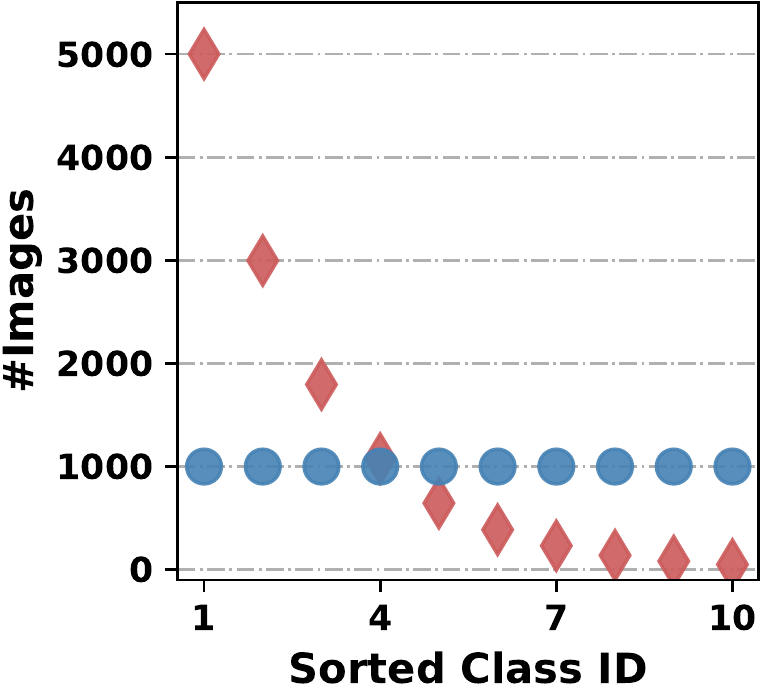}\hspace{0.5mm}
\includegraphics[height=0.295\textwidth]{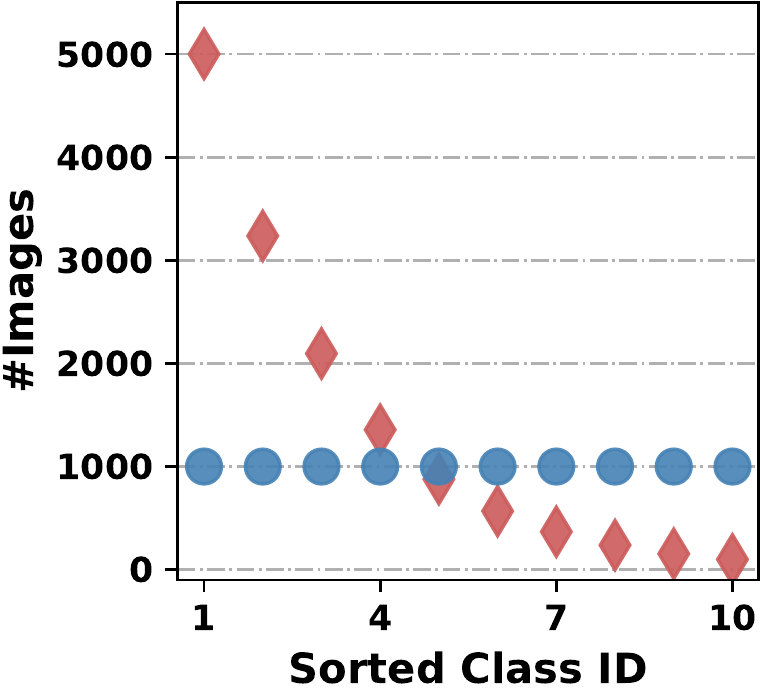}\hspace{0.5mm}
\includegraphics[height=0.295\textwidth]{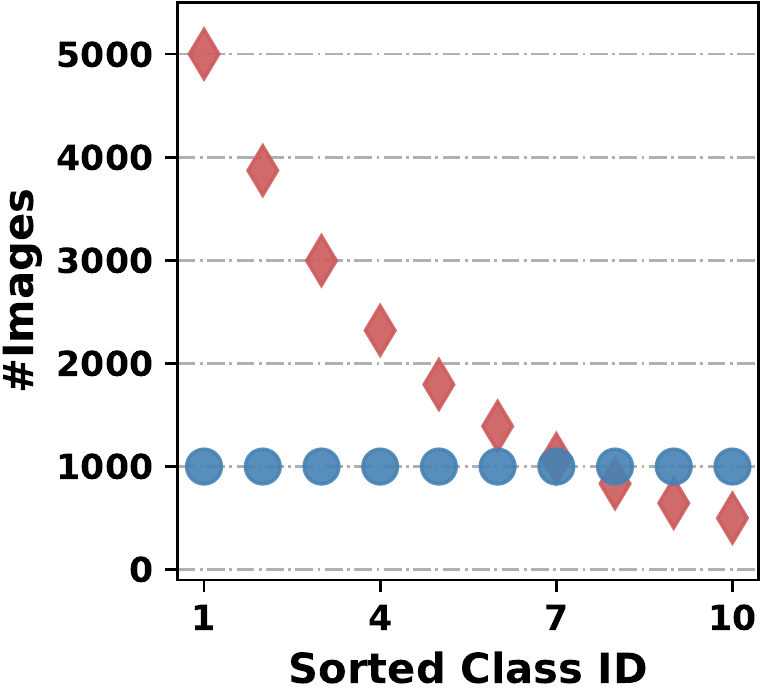}
\caption{Frequency distributions of CIFAR10-LT with imbalance ratio 100, 50, 10 (from left to right).}
\label{fig:lt_distri_cifar10}
\end{minipage}
\\
\begin{minipage}[b]{0.27\textwidth}
\centering
\includegraphics[height=0.78\textwidth]{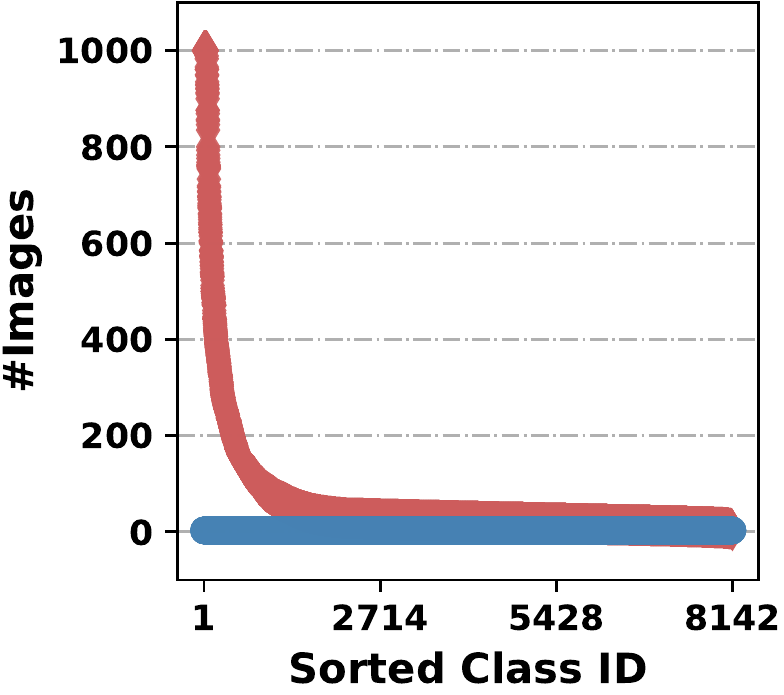}
\caption{Frequency distributions of iNaturalist.}
\label{fig:lt_distri_inat18}
\end{minipage}
\hspace{1mm}
\begin{minipage}[b]{0.7\textwidth}
\includegraphics[height=0.295\textwidth]{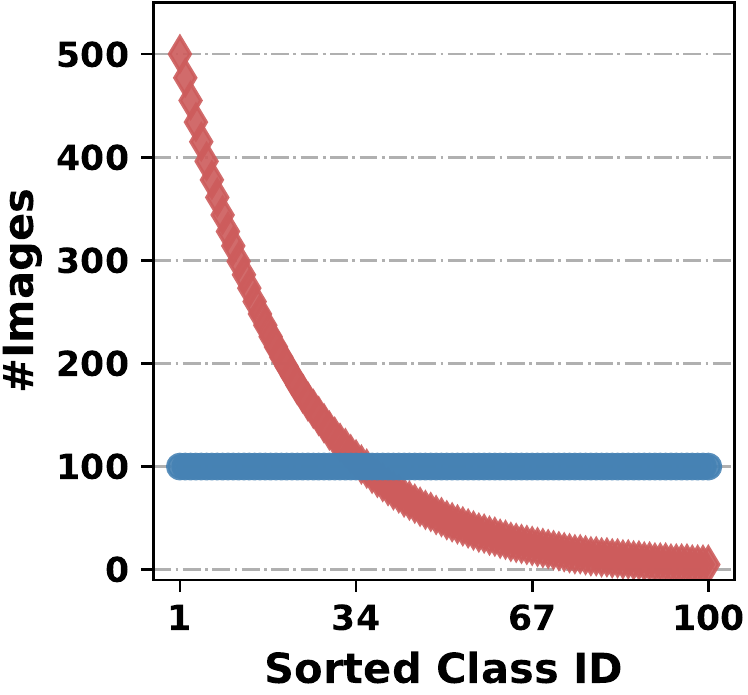}\hspace{0.5mm}
\includegraphics[height=0.295\textwidth]{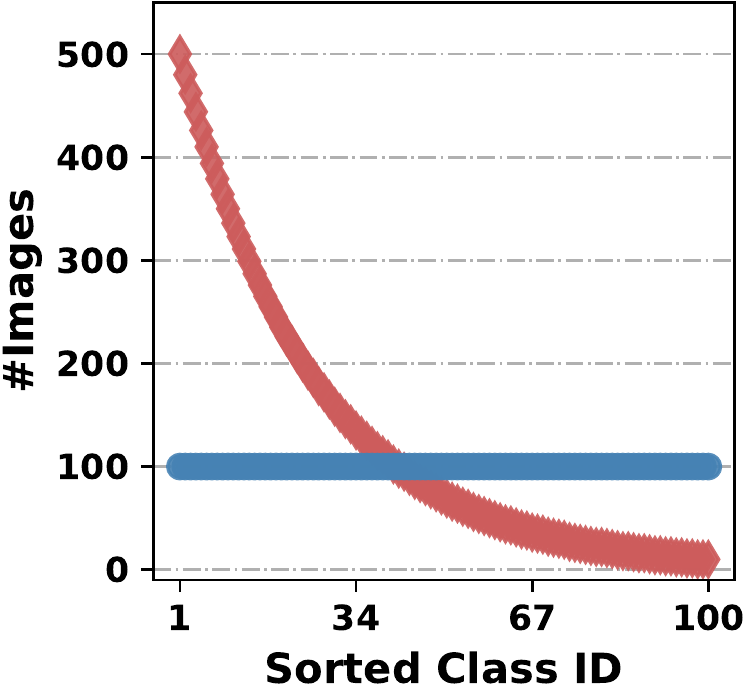}\hspace{0.5mm}
\includegraphics[height=0.295\textwidth]{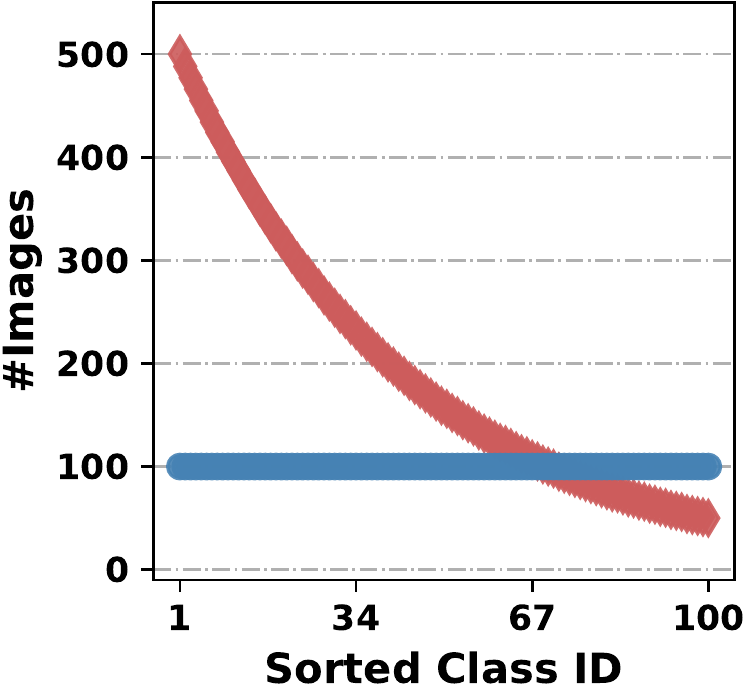}
\caption{Frequency distributions of CIFAR100-LT with imbalance ratio 100, 50, 10 (from left to right).}
\label{fig:lt_distri_cifar100}
\end{minipage}
\end{figure}

For long-tailed recognition, we present the frequency distributions of ImageNet-LT in Fig.~\ref{fig:lt_distri_imagenet}, iNaturalist in Fig.~\ref{fig:lt_distri_inat18}, CIFAR10-LT with different imbalance ratios (the imbalance ratio is defined as the ratio of the number of training samples in the least frequent class and the most frequent class) in Fig.~\ref{fig:lt_distri_cifar10}, and CIFAR100-LT with different imbalance ratios in Fig.~\ref{fig:lt_distri_cifar100}, which also show a severe class imbalance problem.

\section{More Results}
\label{app:mr}

\subsection{Results for Class Incremental Learning}
\label{app:mr_il}

Additionally, we conduct the class incremental learning experiments on CIFAR100 with 20, 10, 5 and 2 incremental steps (5, 10, 20 and 50 classes per incremental step, respectively) with performance listed in Tab.~\ref{tab:cifar100_20}, Tab.~\ref{tab:cifar100_10}, Tab.~\ref{tab:cifar100_5} and Tab.~\ref{tab:cifar100_2} respectively. We find that energy aligning can be generalized to different datasets and settings, achieving improved or competitive results compared to state-of-the-art methods.

\subsection{Results for Long-Tailed Recognition}
\label{app:mr_lt}

We further perform the long-tailed recognition experiments on CIFAR10-LT and CIFAR100-LT with different imbalance ratios with results presented in Tab.~\ref{tab:lt_cifar10_cifar100}, which once again demonstrate the effectiveness of the proposed method.

\section{Combined with Different Classifiers}
\label{app:diff_cls}

\begin{figure}[t]
\centering
\subfloat[cosine classifier, on ImageNet100, for class incremental learning]{
\includegraphics[height=0.2\textwidth]{./fig/il_eff_top5.pdf}
\includegraphics[height=0.2\textwidth]{./fig/il_eff_top1.pdf}
} 
\hspace{8mm}
\subfloat[linear classifier, on ImageNet100, for class incremental learning]{
\includegraphics[height=0.2\textwidth]{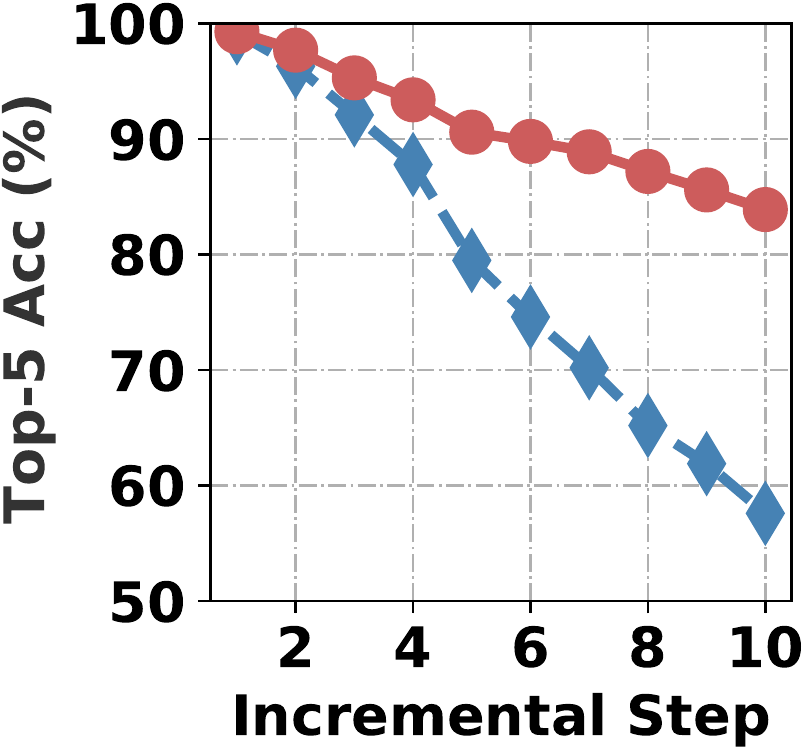}
\includegraphics[height=0.2\textwidth]{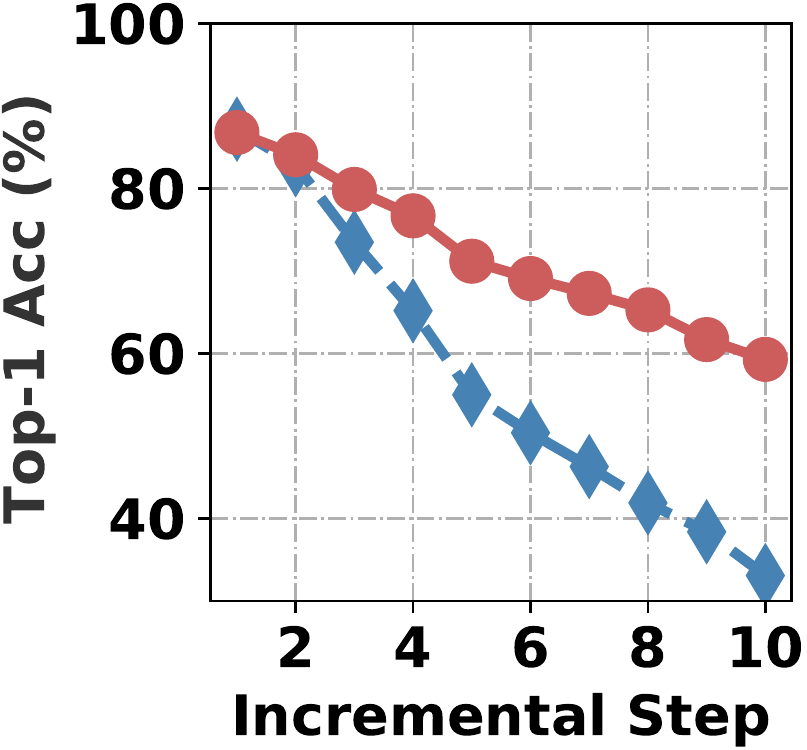}
} 
\\
\subfloat[cosine classifier, on ImageNet1000, for class incremental learning]{
\includegraphics[height=0.2\textwidth]{./fig/il_eff2_top5.pdf}
\includegraphics[height=0.2\textwidth]{./fig/il_eff2_top1.pdf}
} 
\hspace{8mm}
\subfloat[linear classifier, on ImageNet1000, for class incremental learning]{
\includegraphics[height=0.2\textwidth]{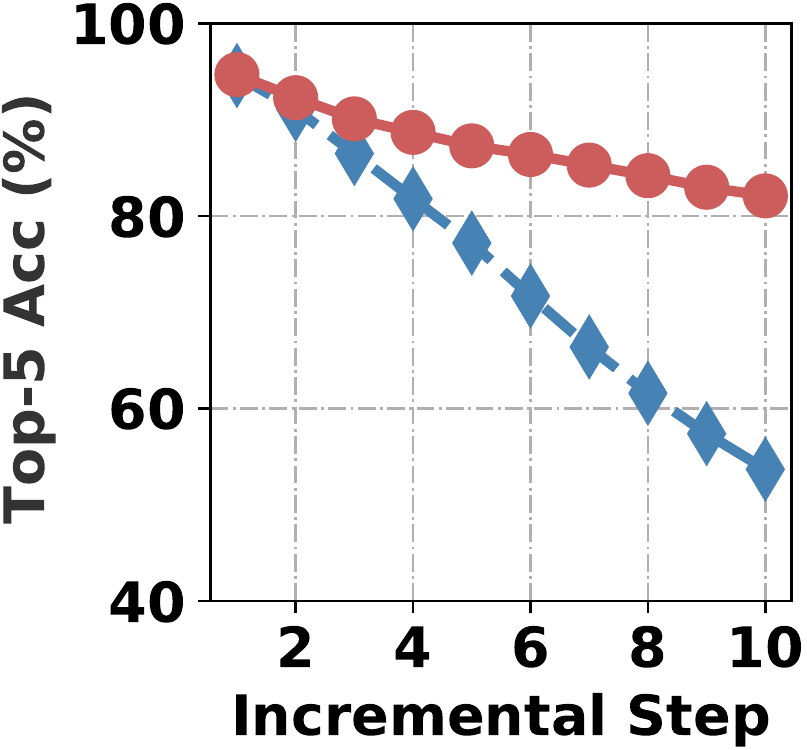}
\includegraphics[height=0.2\textwidth]{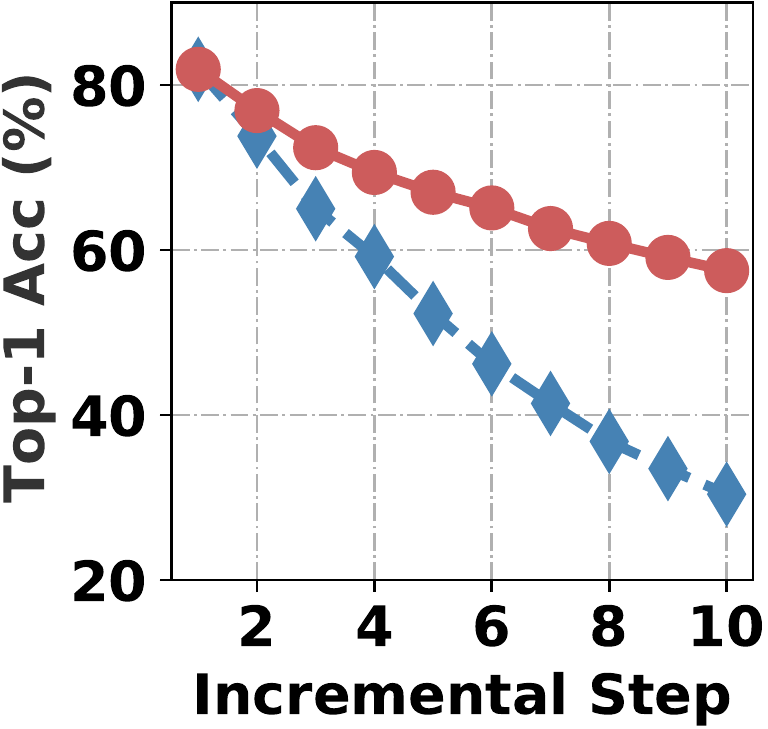}
} 
\\
\subfloat[cosine classifier, on ImageNet-LT, for long-tailed recognition]{\includegraphics[height=0.28\textwidth]{./fig/lt_eff2.pdf}} 
\hspace{8mm}
\subfloat[linear classifier, on ImageNet-LT, for long-tailed recognition]{\includegraphics[height=0.28\textwidth]{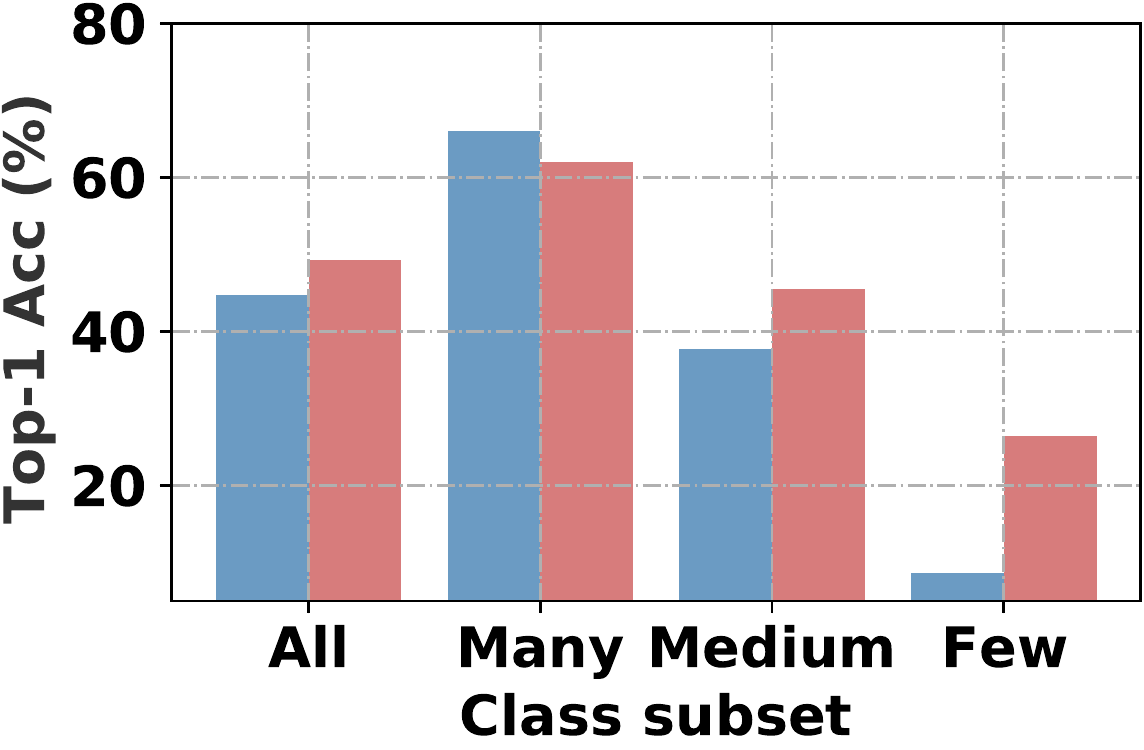}}
\\
\subfloat[cosine classifier, on iNaturalist, for long-tailed recognition]{\includegraphics[height=0.28\textwidth]{./fig/lt_eff.pdf}} 
\hspace{8mm}
\subfloat[linear classifier, on iNaturalist, for long-tailed recognition]{\includegraphics[height=0.28\textwidth]{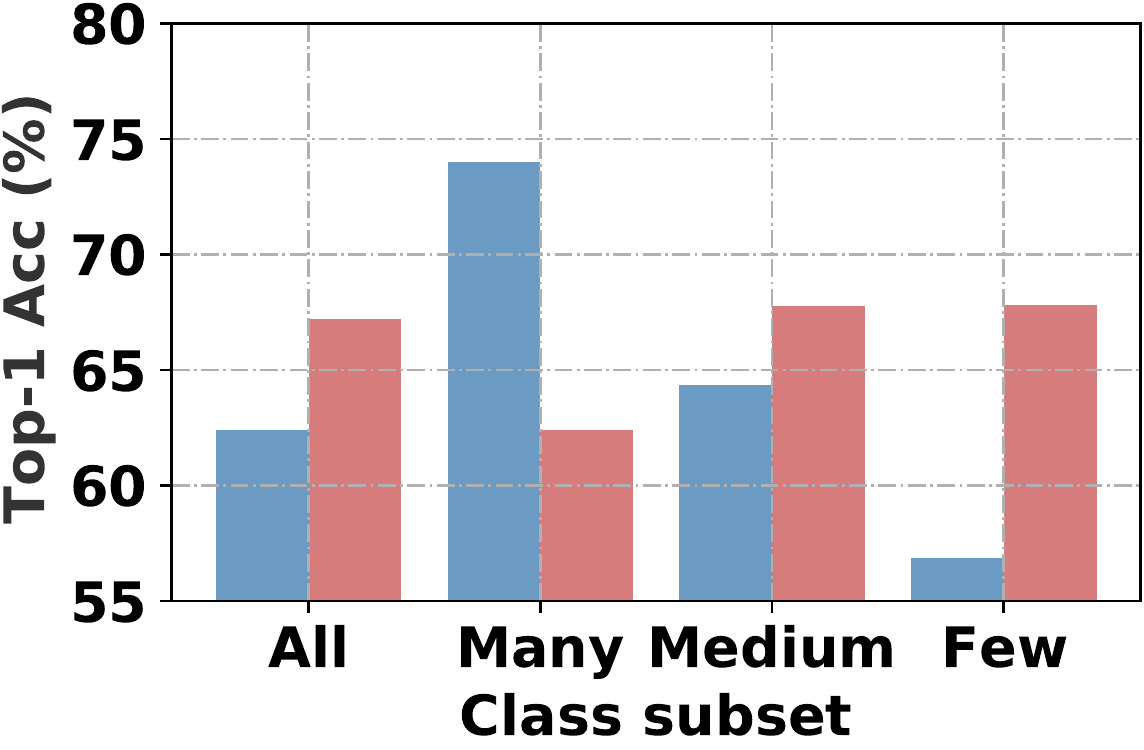}}
\caption{Performance of energy aligning with different classifiers.}
\label{fig:cls}
\end{figure}

We further confirm the robustness of energy aligning with different classifiers through experiments. As shown in Fig.~\ref{fig:cls}, no matter whether using a cosine classifier or a linear classifier, the proposed energy aligning can bring significant improvements. It is worth noting that some previous methods, such as weight aligning~\cite{Zhao_2020_CVPR} and $\tau$-normalized~\cite{DBLP:conf/iclr/KangXRYGFK20}, will no longer benefit the model trained with cosine classifier due to the weights in fully connected (FC) layers has been normalized by the cosine classifier. However, in this work, we show that even with cosine classifier, the trained model is also biased and the proposed energy aligning can still improve the strong baseline by a large margin.

\section{Combined with Different Optimizers}
\label{app:diff_optim}

\begin{figure}[t]
    \centering
    \subfloat[SGD with momentum]{\includegraphics[height=0.28\textwidth]{./fig/lt_eff2.pdf}} \\
    \subfloat[Adam]{\includegraphics[height=0.28\textwidth]{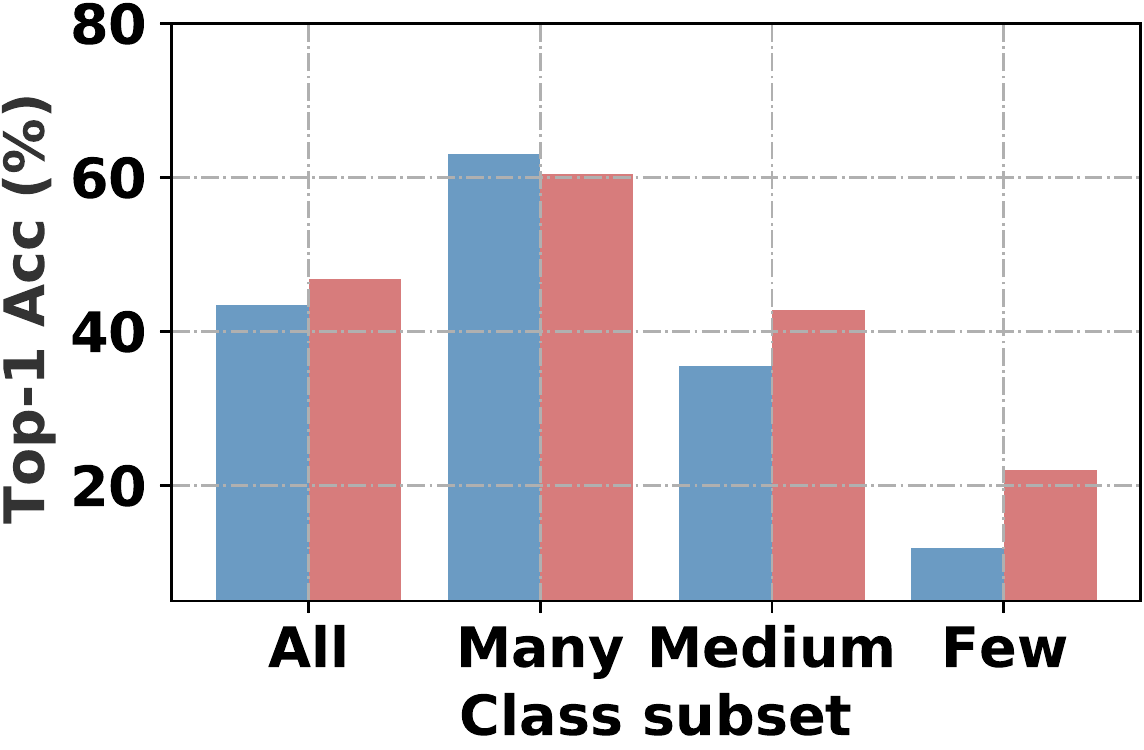}} \\
    \subfloat[SGD without momentum]{\includegraphics[height=0.28\textwidth]{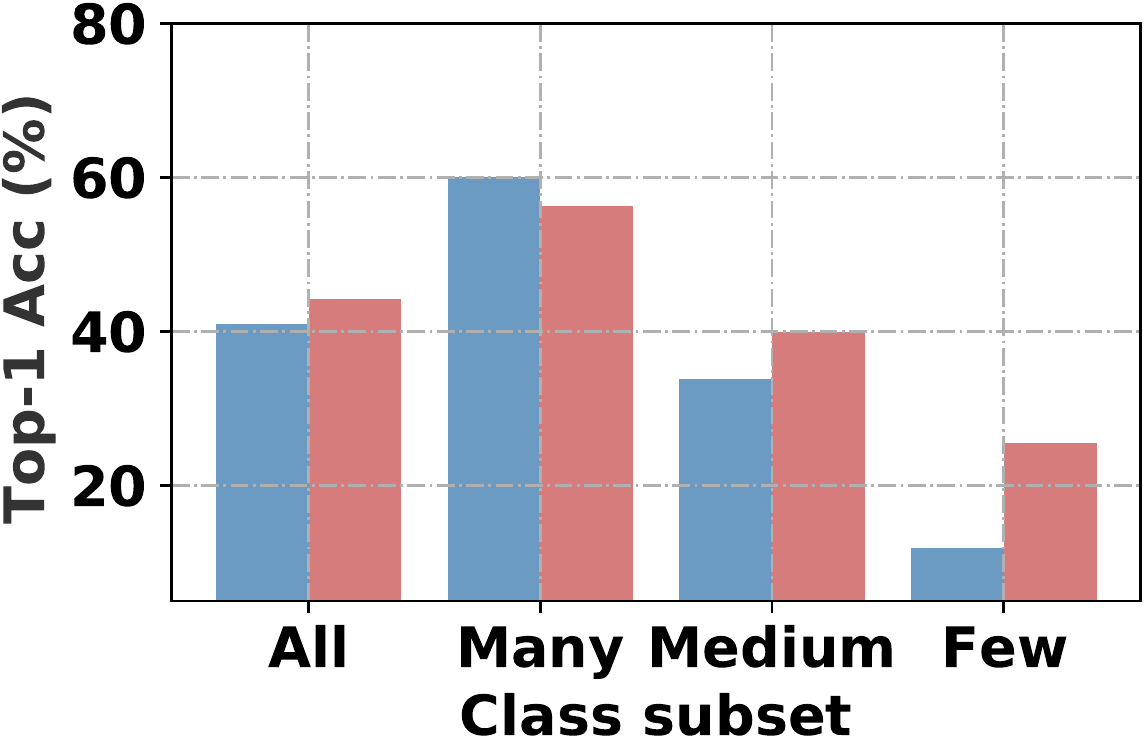}}
    \caption{Performance of energy aligning with different optimizers on ImageNet-LT.}
    \label{fig:optim}
\end{figure}

We also examine the effectiveness of energy aligning with different optimizers on ImageNet-LT. We use SGD with momentum~\cite{sutskever2013importance}, Adam~\cite{kingma2014adam} and SGD without momentum to train models, with statistics presented in Fig.~\ref{fig:optim}. It can be seen that energy aligning achieves improvements on all three models trained with different optimizers. Recent study shows that when training with SGD (with momentum), the norms of weights in FC layers are strongly correlated with class frequency, while with Adam, this phenomenon does not exist~\cite{menon2020long}, so that weight aligning~\cite{Zhao_2020_CVPR} and $\tau$-normalized~\cite{DBLP:conf/iclr/KangXRYGFK20} may fail in such scenarios. However, energy aligning does not suffer this issue. Besides, we also obtain the baseline results when training without momentum (200 epochs), and improve it by a large margin with the help of energy aligning. It can be seen that the model trained without momentum is still heavily biased which can not be fully explained by ~\cite{tang2020long}, whereas can be corrected by energy aligning as well.

\begin{figure}[t]
\begin{minipage}[b]{0.99\textwidth}
\begin{table}[H]
\centering
\caption{Class incremental learning performance (top-1 accuracy \%) on CIFAR100 with 20 incremental steps (5 classes per step). The best results are in bold.}
\begin{tabular}{cc|c|c|c|c|c|c|c|c|c|c}
\toprule
\#step & 1     & 2    & 3    & 4    & 5    & 6    & 7    & 8    & 9    & 10    &  \\
\midrule
LwF~\cite{li2017learning} & 89.4  & 69.1  & 59.7  & 50.9  & 44.6  & 38.9  & 34.9  & 30.6  & 27.7  & 25.7  &  \\
iCaRL~\cite{rebuffi2017icarl} & 89.7  & 82.6  & 77.7  & 74.6  & 70.9  & 68.6  & 66.0  & 63.4  & 61.1  & 59.4  &  \\
EEIL~\cite{castro2018end} & 82.5  & 86.8  & \textbf{84.8} & \textbf{81.0} & \textbf{77.7} & \textbf{74.4} & \textbf{70.6} & \textbf{67.9} & 65.3 & 63.0 &  \\
BiC~\cite{wu2019large} & 95.8  & 90.3  & 80.8  & 75.8  & 73.6  & 71.6  & 67.9  & 65.5  & 62.9  & 61.9  &  \\
WA~\cite{Zhao_2020_CVPR} & 97.6  & 91.6 & 82.3  & 76.5  & 73.9  & 71.6  & 69.6  & 66.3  & 65.2  & 62.4  &  \\
EA (Ours)  & 97.6  & \textbf{92.0}  & 84.5 & 78.2  &  76.7  &  71.8 &  70.4 & \textbf{67.9}  & \textbf{66.3} & \textbf{64.5}  &  \\
\midrule
\#step & 11    & 12    & 13    & 14    & 15    & 16    & 17    & 18    & 19    & 20   & Avg \\
\midrule
LwF~\cite{li2017learning} & 24.0  & 22.0  & 20.0  & 19.1  & 18.3  & 17.1  & 16.3  & 15.7  & 14.9  & 14.3  & 29.7  \\
iCaRL~\cite{rebuffi2017icarl} & 58.0  & 56.3  & 54.9  & 52.9  & 51.1  & 50.0  & 48.0  & 47.1  & 46.0  & 44.9  & 59.7  \\
EEIL~\cite{castro2018end} & 61.3 & \textbf{59.2} & \textbf{57.7} & 55.2  & 53.7  & 51.9  & \textbf{50.6} & 49.4  & 47.9  & 46.8  & \textbf{63.4} \\
BiC~\cite{wu2019large} & 59.3  & 57.3  & 56.2  & \textbf{55.9} & 54.0  & \textbf{52.6} & 49.8  & \textbf{49.6} & \textbf{48.2} & \textbf{47.0} & 62.1  \\
WA~\cite{Zhao_2020_CVPR}  & 61.1  & 58.9  & 56.9  & 55.3  & \textbf{54.5} & 52.0  & 50.1  & 48.0  & 46.8  & 46.0  & 62.6  \\
EA (Ours) & \textbf{62.7} & 58.4  & 56.4 &  55.4  & 53.6 & 51.6  &  50.5 & 47.9  &  46.5 &  45.1 &  63.2 \\
\bottomrule
\end{tabular}%
\label{tab:cifar100_20}%
\end{table}%
\end{minipage}

\begin{minipage}[b]{0.99\textwidth}
\begin{table}[H]
\centering
\caption{Class incremental learning performance (top-1 accuracy \%) on CIFAR100 with 10 incremental steps (10 classes per step). The best results are in bold.}
\begin{tabular}{c|c|c|c|c|c|c|c|c|c|c|c}
\toprule
\#step & 1    & 2    & 3    & 4   & 5   & 6    & 7    & 8    & 9  & 10  & Avg \\
\midrule
LwF~\cite{li2017learning} & 85.4  & 68.9  & 54.9  & 46.3  & 40.6  & 36.6  & 31.4  & 28.6  & 26.0  & 24.6  & 39.7  \\
iCaRL~\cite{rebuffi2017icarl} & 86.0  & 78.6  & 72.6  & 67.4  & 63.7  & 60.6  & 56.9  & 54.3  & 51.4  & 49.1  & 61.6  \\
EEIL~\cite{castro2018end} & 80.2  & 80.9  & 76.1 & 71.3 & 66.2  & 62.5  & 58.9  & 54.8  & 52.2  & 49.5  & 63.6  \\
BiC~\cite{wu2019large} & 90.3  & \textbf{82.2} & 75.2  & 70.2  & 65.5  & 61.3  & 57.7  & 55.2  & 53.7  & 50.2  & 63.5  \\
WA~\cite{Zhao_2020_CVPR}  & 92.1  & 79.7  & 75.6  & 70.3  & 66.4 & \textbf{63.3} & 61.0 & \textbf{57.0} & 54.7 & \textbf{52.4} & 64.5 \\
EA (Ours) & 92.5 & 81.7  &  \textbf{77.0} &  \textbf{71.6}  &  \textbf{68.3}  &  63.0 &  \textbf{61.1}  & 56.6  & \textbf{54.9}  &  \textbf{52.4} &  \textbf{65.2} \\
\bottomrule
\end{tabular}%
\label{tab:cifar100_10}%
\end{table}%
\end{minipage}
\end{figure}

\begin{figure}[t]
\begin{minipage}[b]{0.99\textwidth}
\begin{table}[H]
\centering
\caption{Class incremental learning performance (top-1 accuracy \%) on CIFAR100 with 5 incremental steps (20 classes per step). The best results are in bold.}
\begin{tabular}{c|c|c|c|c|c|c}
\toprule
\#step & 1    & 2    & 3    & 4    & 5   & Avg \\
\midrule
LwF~\cite{li2017learning} & 82.3  & 62.6  & 50.3  & 41.1  & 34.6  & 47.1  \\
iCaRL~\cite{rebuffi2017icarl} & 82.9  & 73.1  & 66.0  & 59.7  & 54.3  & 63.3  \\
EEIL~\cite{castro2018end} & 80.7  & 74.6  & 66.7  & 59.9  & 53.6  & 63.7  \\
BiC~\cite{wu2019large} & 84.0  & 74.7  & 67.9  & 61.3  & 56.7  & 65.1  \\
WA~\cite{Zhao_2020_CVPR}  & 83.5  & 75.5 & 68.7 & \textbf{63.1} & \textbf{59.2} & \textbf{66.6} \\
EA (Ours)  & 82.3 &  \textbf{75.6} & \textbf{69.1} &  62.6 & \textbf{59.2}  & \textbf{66.6} \\
\bottomrule
\end{tabular}%
\label{tab:cifar100_5}%
\end{table}%
\end{minipage}

\begin{minipage}[b]{0.99\textwidth}
\begin{table}[H]
\centering
\caption{Class incremental learning performance (top-1 accuracy \%) on CIFAR100 with 2 incremental steps (50 classes per step). The best results are in bold.}
\begin{tabular}{c|c|c|c}
\toprule
\#step & 1    & 2   & Avg \\
\midrule
LwF~\cite{li2017learning} & 75.7  & 52.6  & 52.6  \\
iCaRL~\cite{rebuffi2017icarl} & 74.9  & 62.0  & 62.0  \\
EEIL~\cite{castro2018end} & 74.1  & 60.8  & 60.8  \\
BiC~\cite{wu2019large} & 76.4  & 64.9  & 64.9  \\
WA~\cite{Zhao_2020_CVPR}  & 78.0  & 65.1 & 65.1 \\
EA (Ours)  & 76.9 & \textbf{65.2}  & \textbf{65.2} \\
\bottomrule
\end{tabular}%
\label{tab:cifar100_2}%
\end{table}%
\end{minipage}

\begin{minipage}[b]{0.99\textwidth}
\begin{table}[H]
\centering
\caption{Top-1 accuracy (\%) on CIFAR100-LT and CIFAR10-LT with different imbalance ratios. The best results are in bold.}
\begin{tabular}{l|c|c|c|c|c|c}
\toprule
Dataset & \multicolumn{3}{c|}{CIFAR100-LT} & \multicolumn{3}{c}{CIFAR10-LT} \\
\midrule
Imbalance ratio & 100   & 50    & 10    & 100   & 50    & 10 \\
\midrule
Focal~\cite{lin2017focal} & 38.4 & 44.3 & 55.8 & 70.4 & 76.7 & 86.7 \\
Mixup~\cite{zhang2017mixup} & 39.5 & 45.0 & 58.0 & 73.1 & 77.8 & 87.1 \\
Manifold Mixup~\cite{verma2019manifold} & 38.3 & 43.1 & 56.6 & 73.0 & 78.0 & 87.0 \\
CE-DEW~\cite{cao2019learning} & 41.5 & 45.3 & 58.1 & 76.3 & 80.0 & 87.6 \\
CE-DRS~\cite{cao2019learning} & 41.6 & 45.5 & 58.1 & 75.6 & 79.8 & 87.4 \\
CB-Focal~\cite{cui2019class} & 39.6  & 45.2 & 58.0 & 74.6 & 79.3 & 87.1 \\
LDAM-DRW~\cite{cao2019learning} & 42.0 & 46.6 & 58.7 & 77.0 & 81.0 & 88.2 \\
BBN~\cite{zhou2020bbn} & 42.6 & 47.0 & 59.1 & 79.8 & 82.2 & 88.3 \\
De-confound~\cite{tang2020long} & 40.5  & 46.2  & 58.9  & 71.7  & 77.8  & 86.8 \\
De-confound-TDE~\cite{tang2020long} & 44.1  & 50.3  & 59.6  & \textbf{80.6}  & \textbf{83.6} &  \textbf{88.5} \\
LA-post-hoc~\cite{menon2020long} & 41.8  & -- & -- & 77.4  & -- & -- \\
LA-loss~\cite{menon2020long} & 43.9  & -- & -- & 77.7  & -- & --  \\
LA-loss + LDAM~\cite{menon2020long} & 44.1  & -- & -- & 77.6  & -- & --  \\
EA (Ours) & \textbf{45.3} & \textbf{50.6} & \textbf{60.5} & 80.5  & 82.7  &  88.2 \\
\bottomrule
\end{tabular}%
\label{tab:lt_cifar10_cifar100}%
\end{table}%
\end{minipage}
\end{figure}

\end{document}